\documentclass[preprint,review,12pt]{elsarticle}

\usepackage{amssymb}
\usepackage{graphicx}
\usepackage{amsmath}
\usepackage{amssymb}
\usepackage{amstext}
\usepackage{amsthm}
\usepackage{color}	
\usepackage{url}
\usepackage{calc}
\usepackage{hyperref}
\usepackage{algorithm}
\usepackage{algorithmic}
\usepackage{arydshln}
\usepackage{booktabs}
\usepackage{subcaption}
\usepackage{mdframed}
\usepackage{xcolor,lipsum}
\biboptions{sort&compress}
\usepackage{float}
\usepackage{multirow}
\usepackage{adjustbox}
\usepackage{pslatex}
\usepackage{colortbl}

\journal{Image and Vision Computing}

\begin{document}

\begin{frontmatter}

\title{Person Re-ID through Unsupervised Hypergraph \\ Rank Selection and Fusion}

\author{Lucas Pascotti Valem, Daniel Carlos Guimarães Pedronette}
\address{Department of Statistics, Applied Mathematics and Computing (DEMAC),\\  S\~{a}o Paulo State University (UNESP), Rio Claro, Brazil}

\begin{abstract}

Person Re-ID has been gaining a lot of attention and nowadays is of fundamental importance in many camera surveillance applications. The task consists of identifying individuals across multiple cameras that have no overlapping views. Most of the approaches require labeled data, which is not always available, given the huge amount of demanded data and the difficulty of manually assigning a class for each individual. Recently, studies have shown that re-ranking methods are capable of achieving significant gains, especially in the absence of labeled data. Besides that, the fusion of feature extractors and multiple-source training is another promising research direction not extensively exploited. We aim to fill this gap through a manifold rank aggregation approach capable of exploiting the complementarity of different person Re-ID rankers. In this work, we perform a completely unsupervised selection and fusion of diverse ranked lists obtained from multiple and diverse feature extractors. Among the contributions, this work proposes a query performance prediction measure that models the relationship among images considering a hypergraph structure and does not require the use of any labeled data. Expressive gains were obtained in four datasets commonly used for person Re-ID. We achieved results competitive to the state-of-the-art in most of the scenarios.

\end{abstract}


\end{frontmatter}

\section{Introduction}
\label{sec:introduction}

Person Re-ID is of critical importance in the majority of modern security and surveillance applications~\cite{paperReIDApp1, paperReIDApp2, paperReIDApp3, paperReIDSurvey2016}.
The task consists in, given a query image of one person, to identify the same individual across different cameras which have no overlapping views.
There are many difficulties for Re-ID retrieval~\cite{deepSurvey2020}, among them:
\textit{(i)} different viewpoints,
\textit{(ii)} possible low-image resolutions,
\textit{(iii)} illumination changes,
\textit{(iv)} occlusions,
\textit{(v)} difficulty of manually labeling data for training,
\textit{(vi)} large amount of data to be processed.
The challenge of improving the effectiveness of these systems, especially in open-world scenarios, has attracted a lot of research efforts from the scientific community~\cite{airport, paperReIDSurvey2016, deepSurvey2020}.

Initially, person Re-ID retrieval systems were mainly based on the use of hand-crafted feature representations~\cite{paperSalientColorFeature, paperELF, paperGBICOV, paperGOG, paperLOMO, paperWHOS, market1501, paperHLBP, cuhk03}. Besides that, other strategies commonly used to generic image retrieval, like bag of visual words~\cite{paperSDC, market1501}, have also been employed.
Aiming at further improving the quality of the results, rather than using traditional distance measures, researchers have proposed metric learning approaches for Re-ID~\cite{paperLOMO, paperLargeScaleML, paperModelFeatDistClass, paperKernelizedReidML, zhang2016learning}, most of them based on supervised models.
In~\cite{airport}, an extensive evaluation of multiple combinations of feature extractors and metric learning approaches is discussed.

Due to the significant impact of deep learning on common image retrieval and machine learning~\cite{deepSurvey2020}, the Convolutional Neural Networks (CNN) also have gained a lot of attention and have  been widely employed to solve Re-ID tasks in the recent years~\cite{deepSurvey2020, paperOSNET, paperOSNET-IBN-AIN, paperMLFN, paperHACNN, paperPCB, paperLargestMS, paperEANET, paperATNET, paperAANET}.
Among the works, existing networks have been trained for Re-ID~\cite{paperRESNET, paperMobileNetv2}, and some new architectures have been proposed solely focusing on Re-ID~\cite{paperOSNET, paperOSNET-IBN-AIN, paperMLFN, paperHACNN, paperPCB}.

Despite the success of deep learning, one of the main difficulties resides in the lack of large publicly available Re-ID datasets for training~\cite{deepSurvey2020}, mainly because manually labeling images is a very difficulty task, especially in open-world scenarios~\cite{airport}, where the datasets may increase dynamically.
In order to mitigate this problem, many authors have proposed multi-source training, which consists in joining multiple available datasets for training~\cite{paperEMTL, paperCAMEL, paperLargestMS}, usually increasing the network capacity for generalization.

However, despite several advances in multi-source approaches, it is inevitable the need for intensive manual annotation for obtaining training data.
In practice, the demand for extensive training data restricts the generalization and scalability of supervised approaches, especially on person Re-ID tasks, which are not only resource-intensive to acquire identity annotation but also impractical for large-scale data~\cite{PaperCrossSimTIP20,PaperUnsReIDSimLenCVPR20}.

Usually, there are two main steps associated to Re-ID labeling process~\cite{paperIntraCam2021, deepSurvey2020, paperIntraCam2019}: \textit{(i)} intra-camera annotation that requires to compare a person with all the other unlabeled persons in a single camera with multiple views; and \textit{(ii)} inter-camera that requires to match a person across different cameras with multiple views.
Let $P$ be the number of persons and $S$ the number of camera views.
The intra-camera annotation complexity is $O(S \times P^2)$ and inter-camera annotation complexity ranges from $O(S \times P^2)$ and $O(S^2 \times P^2)$.
The worst case of inter-camera occurs because not all persons appear in every camera view in the majority of cases, which makes the association requiring to repeat for all $V$ camera views~\cite{paperIntraCam2021, paperIntraCam2019}.
Commonly, inter-camera association significantly increases standard annotation costs.
There are approaches that mitigate these issues; among them there is the Intra-Camera Supervised (ICS)~\cite{paperIntraCam2021, paperIntraCam2019}, Multi-Task Multi-Label (MATE)~\cite{paperIntraCam2021, paperIntraCam2019}, and Cross-camera Feature Prediction~\cite{paperCrossCamSup2021}, which were recently proposed.

Due to the challenge of obtaining large amounts of strongly labeled data, semi-supervised methods have been employed, which is a typical strategy for supervision minimization.
Based on information learned from a small set of labeled data, the idea is to generate labels from unlabeled training data.
Some research has been made in this direction~\cite{paperSemiSup2013, paperSemiSup2014, paperOpenSetReid2016, paperSemiSup2019}.
However, these methods often suffer from performance degradation and often require a large proportion of expensive cross-view pairwise labeling~\cite{paperIntraCam2021}.

There are also weakly supervised strategies that replace accurate labels with inaccurate annotations.
In~\cite{weaklyreid2019}, the authors proposed the idea of obtaining multiple bounding boxes of the same person from untrimmed videos. This is done by training a deep learning model capable of extracting multiple bounding boxes of the same person in a video.
Recently,~\cite{wang2020weakly} proposed to replace image-level annotations with bag-level annotations.
Weakly supervised Re-ID is very challenging, since it is rather difficult to model the considerable variances across camera views (e.g. occlusion and illumination) without using strong label data~\cite{weaklyreid2019, paperIntraCam2021}.

As an alternative solution, unsupervised approaches~\cite{paperARN, paperECN, paperMAR, paperTAUDL, paperUTAL, paperUnsCrossDom2019, paperDECAMEL,PaperCrossSimTIP20,PaperUnsReIDSimLenCVPR20} have been attracting a lot of attention from the research community, especially because, once labeled data are not required, the methods become more suitable for real-world scenarios.
In a promising research direction, to address the lack of labels issue, there are works proposed to post-process person Re-ID results by analyzing similarity relationships encoded in the datasets.
Several authors have proposed unsupervised re-ranking approaches for Re-ID~\cite{cuhk03_new_protocol, paperCNearNeighborReRank, paperRRReID, paperLeng2014, paperQueryBasedAdaptativeRR, paperReIDRerankAnalysis, paperMang2015, paperDiscInfAnRR, paperGuo2018}.
In~\cite{cuhk03_new_protocol},  the original distances among images are improved by calculating Jaccard correlation scores of each ranked list.
Various approaches exploit the reciprocal neighborhood information and other co-occurrence indexes aiming at improving the ranking results.

Another strategy commonly employed on generic unsupervised image retrieval and few exploited on Re-ID tasks consists in fusion approaches~\cite{paperPirasFusion17}. 
 In general, the  both broad categories of fusion have been successfully used in generic image retrieval: \textit{(i)} early fusion, which combines the feature vectors; and \textit{(ii)} late fusion, which usually combines ranked lists.
Significant results have been achieved based on fusion of  different ranked lists and features~\cite{paperUSRAF, fusionRetrieval}, with the purpose of obtaining more effective results by exploiting the complementarity of each input.

On person Re-ID tasks, some authors have proposed late fusion strategies based on rank aggregation~\cite{paperRankAggReID2016, paperQueryAdaptiveFusion}.
In~\cite{paperQueryAdaptiveFusion}, the aggregation is performed by attributing weights for each query of each ranker, but no pre-selection of features is performed, and all the features are used as input for the fusion step.
There are also early fusion approaches for Re-ID, as~\cite{paperHypergraphFeatureFusion17},  that propose a supervised multi-hypergraph fusion model for early fusion of feature extractors.
It learns a hypergraph for each feature  through a star expansion strategy and they are fused according to weights that the method has learned from training.
In~\cite{paperHypergraphReID19}, a hypergraph structure is used with a deep learning model to improve the performance of the acquired features for Re-ID.

Although fusing different features can represent a significant advantage due to extra information available, how to choose what features to fuse can be a challenging task.
Even for supervised approaches, selecting high-effective combinations of visual features remains a complex task, since it is necessary to consider
various aspects, such as diversity and complementarity of  results.
Therefore, selecting features in an unsupervised way, without
any labeled data is even more challenging, since no information about the effectiveness of individual visual features is available.

This paper addresses the challenging task of unsupervised selection and fusion of different features for more effective person re-identification.
We propose a novel Hypergraph Rank Selection and Fusion (HRSF) framework, which combines an unsupervised rank-based formulation for feature selection~\cite{paperUSRAF} with a robust hypergraph model~\cite{paperLHRR} for query performance prediction and rank aggregation based on manifold learning.
Among our main contributions, we can highlight:

\begin{itemize}

    \item The proposed HRSF framework uses a rank-based late fusion model, suitable for selection and fusion of a broad diversity of features. 
    The selection is performed by exploiting an unsupervised measure for query performance prediction; 

    \item A hypergraph rank-based formulation is used to encode the high-order relationship among images. This strategy is exploited for both selection and fusion tasks. Hypergraph models were little exploited in Re-ID literature~\cite{paperHypergraphFeatureFusion17, paperHypergraphReID19};

    \item The proposed technique is able to learn representations through the hypergraph structure that encodes multiple features from different rankers. The manifold learning based on a hypergraph model~\cite{paperLHRR} allows effective fusion and final ranking. In addition, our approach innovates by fusing different feature extractors trained on different datasets;

    \item Different from most of fusion approaches for Re-ID which often consider ad hoc selections or combine all the features of the input set~\cite{paperRankAggReID2016, paperQueryAdaptiveFusion, paperHypergraphReID19}, our approach is capable of dealing with various features in a completely unsupervised scenario, selecting combinations in a very large search space.
    To the best of our knowledge, this is the first work which performs an explicit selection and subsequently fusion of features on person Re-ID tasks in a completely unsupervised way.
\end{itemize}

In fact, the proposed Hypergraph Rank Selection and Fusion (HRSF) approach is based on both~\cite{paperUSRAF} and~\cite{paperLHRR}.
However, while there are some aspects in common, also there are crucial differences. Among them, we can mention:
\begin{itemize}
    \item \textbf{The problem of selecting and fusing rankers in unsupervised scenarios is very challenging. Hence, the use of an effective algorithm for selection is fundamental}. While USRF~\cite{paperUSRAF} employs traditional query performance prediction approaches (Authority and Reciprocal scores), we propose a new measure named \textit{Hypergraph Query Performance Prediction} (HQPP);
    \item Originally, USRF~\cite{paperUSRAF} uses the CPRR~\cite{paperCPRR} as a method for fusion tasks. Differently, the proposed HRSF uses the LHRR~\cite{paperLHRR} method, which in combination with the HQPP, makes \textbf{both the selection and fusion based on hypergraph structures.} The LHRR~\cite{paperLHRR} method is also more robust than the CPRR~\cite{paperCPRR}, achieving superior retrieval results in most datasets;
    \item Both~\cite{paperUSRAF} and~\cite{paperLHRR} were originally proposed and evaluated only on  general purpose image retrieval scenarios. The proposed HRSF is employed and validated for person Re-ID tasks;
    \item In unsupervised scenarios, the optimal neighborhood size ($k$ parameters) can be challenging to define. The experiments revealed that HRSF is \textbf{more robust than HQPP to different neighborhood sizes}, leading to the most effective results in the majority of scenarios.
\end{itemize}

A wide experimental evaluation was conducted on 4 different datasets with sizes ranging from 14,097 to 39,902 images. Up to 28 different rankers where considered in each case, resulting in millions of possible combinations. 
Experiments indicated that our approach was capable of selecting and fusing the rankers, achieving high-effective results superior to all rankers in isolation and competitive to state-of-the-art, when  more than 20 recent Re-ID approaches are considered.

The remainder of this paper is organized as follows. Section~\ref{secProbDef} discusses a formal definition of problem setting.
Section~\ref{secProposedAproach} presents the proposed rank selection and fusion for Re-ID. 
Section~\ref{secExpEval} discusses the conducted experimental evaluation and, finally, Section~\ref{secConc} presents our conclusions.

\section{Problem Formulation}
\label{secProbDef}

This section presents the formal definition of the rank model used in this work.
Let $\mathcal{C}$=$\{x_{1},$ $x_{2},$ $\dots,x_{N}\}$ be an image collection, where $N$ denotes the collection size.
Let us consider a retrieval task where, given a query image, returns a list of images from the collection $\mathcal{C}$. 

Formally, given a query image $x_q$, a ranker denoted by $R_j$ computes a ranked list $\tau_{q}$=$(x_{1}$, $x_{2}$, $\dots$, $x_{k})$ in response to the query.
The ranked list $\tau_{q}$ can be defined as a permutation of the $k$-neighborhood set $\mathcal{N}(q,k)$, which contains the $k$ most similar images to image $x_q$ in the collection $\mathcal{C}$.
The permutation $\tau_q$ is a bijection from the set $\mathcal{N}(q,k)$ onto the set $[k]=\{1,2,\dots,k\}$.
The $\tau_q (i)$ notation denotes the position of image $x_i$ in the ranked list $\tau_q$.

The ranker $R_j$ can be defined based on diverse approaches, including feature extraction or learning methods. In this paper, a feature-based approach is considered, defining  $R$ as a tuple $(\epsilon,\rho)$, where $\epsilon:$ $\mathcal{C}$ $\rightarrow$ $\mathbb{R}^{d}$ is a function that extracts a feature vector $v_{x}$ from an image $x \in \mathcal{C}$; and $d$: $\mathbb{R}^{d} \times \mathbb{R}^{d} \rightarrow \mathbb{R}$ is a distance function that computes the distance between two images according to their corresponding feature vectors.
Formally, the distance between two images $x_{i}, x_{j}$ is defined by $\rho$($\epsilon(x_{i})$, $\epsilon(x_{j})$). 
The notation $\rho(i,j)$ is used for readability purposes.

A ranked list can be computed by sorting images in a crescent order of distance. In terms of ranking positions we can say that, if image $x_{i}$ is ranked before image $x_{j}$ in the ranked list of image $x_q$, that is, $\tau_q(i) < \tau_q(j)$, then $\rho(q, i)$ $\leq$  $\rho(q, j)$.
Taking every image in the collection as a query image $x_{q}$, a set of ranked lists $\mathcal{T}$ = $\{\tau_{1}, \tau_{2},$ $\dots,$  $\tau_{N}\}$ can be obtained.

Different features and distance functions give rises to different rankers which, in turn, produce distinct ranked lists. 
Let $\mathcal{R}=$ $\{R_1$, $R_2,\dots$, $R_m\}$ be a set of rankers and $R_j \in \mathcal{R}$, we denote by $\mathcal{T}_j$ the set of ranked lists produced by $R_j$. A ranked list computed by the ranker $R_j$ in response to a query $x_q$ is denoted by $\tau_{j,q}$.

Our proposed method (HRSF) intends to select the most effective rankers from $\mathcal{R}$ based on their respective set of ranked lists, without the need of any labeled data.
The selection function ($f_s$) is formally defined by Equation~(\ref{eq:f_selection}).

\vspace{-1mm}
\begin{equation}
\label{eq:f_selection}
\mathfrak{X}^{*}_{n} = f_s (\mathcal{T}_1, \mathcal{T}_2, \mathcal{T}_3, \dots, \mathcal{T}_m ),
\end{equation}
\vspace{-2mm}

\noindent where $\mathfrak{X}^{*}_{n}$ is the set of selected rankers, whose cardinality is $n$, such that $|\mathfrak{X}^{*}_{n}|=n$.

With the objective of facilitating the understanding of the reader, Table~\ref{tab:symbols} presents the main symbols used in this paper.

\begin{table}[ht!]
\centering
\caption{Table of symbols.}
\label{tab:symbols}
\resizebox{.89\textwidth}{!} {
\begin{tabular}{|c|c|l|}
\hline
 \textbf{Type} & \textbf{Symbol} & \textbf{Description} \\
  \hline
  \multirow{6}{*}{\textbf{Retrieval}} & $\mathcal{C}$ & Image collection. \\
  \multirow{6}{*}{\textbf{Model}} & $N$ & Image collection size. \\
   & $x_i$ & Image of index $i$. \\
   & $\mathcal{N}(q,k)$ & Neighborhood set for a query image $x_q$ of size $k$. \\
   & $\tau_{q}$ & Ranked list for the query image $x_q$. \\
   & $\tau_{q}(j)$ & Position of the image $x_j$ in the ranked list of the image $x_q$. \\
   & $L$ & Size of the ranked lists. \\
   & $\mathcal{T}$ & Set of ranked lists for all the images in the dataset. \\
  \hline
  \multirow{15}{*}{\textbf{Selection}} & $f_s$ & Function for ranker selection. \\ 
  \multirow{15}{*}{\textbf{Model}} & $R_{i}$ & Ranker of index $i$. \\ 
  & $\tau_{i,q}$ & Ranked list of the image $x_q$ computed by the ranker $x_i$. \\ 
  & $\mathcal{T}_i$ & Set of ranked lists produced by the ranker $R_{i}$. \\ 
  & $\mathcal{R}$ &  Set of rankers. \\ 
  & $m$ & Size of the set $\mathfrak{R}$. \\ 
  & $\mathfrak{X}_{n}$ &  Candidate combination composed by $n$ rankers. \\ 
  & $\mathfrak{X}^{*}_{n}$ &  Selected combination composed by $n$ rankers. \\ 
  & $\mathfrak{X}^{*}$ &  Selected combination among all sizes. \\ 
  & $n$ & Size of a combination. \\
  & $w_{p}$ & Selection measure for pairs of rankers. \\ 
  & $\beta$ & Weight or relevance of the correlation. \\ 
  & $k$ & Neighborhood size. \\ 
  & $\gamma$ & Effectiveness estimation measure (HQPP). \\
  & $\lambda$ & Correlation measure (RBO). \\
  & $\alpha$ & Constant used in RBO correlation measure. \\
  \hline
  \multirow{6}{*}{\textbf{Hypergraph}} & $V$ & Set of vertexes. \\ 
  \multirow{6}{*}{\textbf{Model}} & $v_{i}$ & Vertex of index $i$. \\
   & $E$ & Set of hyperedges. \\
   & $e_{i}$ & Hyperedge of index $i$. \\
   & $h(e_i, v_j)$ & Reliance of vertex $v_{j}$ to belong to a hyperedge $e_{i}$. \\
   & $r(e_i,v_j)$ & Density of ranking references to $v_j$ in ranking of $x_i$ and its neighbors. \\
   & $\mathbf{H_{G}}$ & Hypergraph model. \\
   & $\mathbf{H}$ & Incidence matrix. \\
   & $\eta_r(i,x)$ & Function that assigns a weight to image $x$ according to its position in $\tau_i$. \\
   & $\eta_f$ & Fused affinity measure used for rank aggregation. \\
  \hline
\end{tabular}
}
\end{table}

\section{Unsupervised Hypergraph Rank Selection and Fusion}
\label{secProposedAproach}

This paper proposes a framework named Hypergraph Rank Selection and Fusion (HRSF) for unsupervised person Re-ID tasks.
Our model is inspired by a recent approach~\cite{paperUSRAF}  proposed for rank selection and fusion on general image retrieval tasks. 
The method is based on effectiveness estimations and correlation among features computed by a rank-based analysis.
In~\cite{paperUSRAF}, reciprocal references are exploited for effectiveness estimation and feature selection, while rank correlation measures are used for analyzing complementary and diversity aspects. The selected features are fused through a rank-based similarity learning method~\cite{paperCPRR}.

The proposed approach differs from previous work~\cite{paperUSRAF} on four main aspects:
(\textit{i}) the unsupervised measure used to estimate the quality of individual features, which is based on hypergraph structures; (\textit{ii}) a more robust method to fuse the selected features; (\textit{iii}) the proposed approach is evaluated and validated for person Re-ID and; (\textit{iv}) it is more robust to different neighborhood sizes, leading to the most effective results in the majority of scenarios.
We innovate by employing a robust hypergraph model for both tasks: query performance prediction and rank fusion.
A recent manifold learning approach based on a rank-based hypergraph formulation~\cite{paperLHRR} is exploited.

The hyperedges weights, used for estimating the confidence of hyperedge associations, are exploited in our approach  to predict the effectiveness of the different person Re-ID rankers in an unsupervised fashion.
Additionally, we use a manifold learning algorithm for fusion tasks, the Log-based Hypergraph of Ranking References (LHRR)~\cite{paperLHRR}.
This keeps our approach completely unsupervised, and more robust, mostly based on  hypergraph structures.

\begin{figure}[!ht]
    \centering
    \includegraphics[width=.87\textwidth]{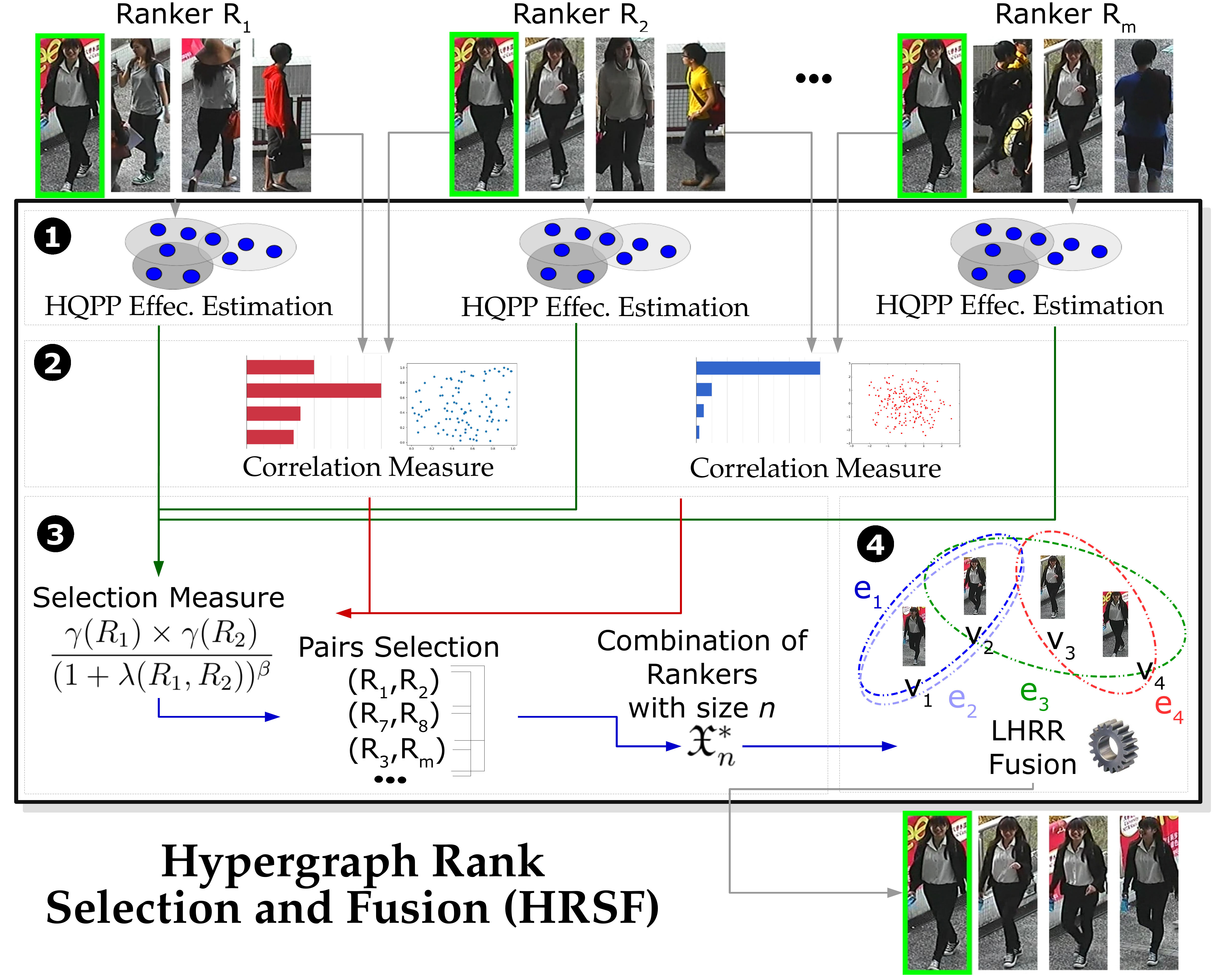}
    \caption{Overview of the HRSF proposed approach.}
    \label{fig:hrsf_diagram}
\end{figure}

Figure~\ref{fig:hrsf_diagram} presents an overview of the proposed approach, where the main steps are illustrated and numerated.
Given a set of different rankers provided by diverse features extractors and distance measures, (1) a hypergraph estimation measure is employed in order to predict the performance of each ranker without using data labels.
In (2), a correlation measure is applied for each pair of rankers.
The computed measures are used in the equation presented in the step (3), which computes the equation for each combination.
The rankers selected in stage (3) are fused in stage (4), which uses LHRR~\cite{paperLHRR} for rank aggregation.

\subsection{Unsupervised Ranker Selection}

Given a set of available rankers for person Re-ID and no labeled data, we aim to select a combination which produces the most effective results.
The selection measure proposed on~\cite{paperUSRAF} relies on the idea that rankers can be analyzed in pairs using an effectiveness estimator and a correlation measure.
It consists in attributing a weight to each pair ($w_{p}$), in such way that, the ones composed by the most effective rankers and the highest/lowest correlated ones should receive a higher score.
This is presented in Equation~(\ref{eq:selection_eq}).

\begin{equation}\label{eq:selection_eq}
w_{p}(\{R_1,R_2\}) = \dfrac{\gamma(R_1) \times \gamma(R_2)}{(1 + \lambda(R_{1},R_{2}))^{\beta}},
\end{equation}

\noindent where $\gamma$ and $\lambda$ are used to measure the effectiveness and correlation of rankers, respectively.
The $\gamma$ corresponds to our proposed Hypergraph Query Performance Prediction (HQPP), which is described in Section~\ref{secHQPP}.
While the effectiveness can be individually estimated for each ranker, the correlation measure is applied to pairs.
The exponent $\beta$ can be employed to decide if the selection favors the most correlated or diverse rankers.
We adopt $\beta=-1$, since in~\cite{paperUSRAF} it was used for scenarios with a higher number of features.

Therefore, in our approach, a ranker pair is selected based on the score obtained by $w_p$.
All ranker pairs are ranked according to $w_p$ (in descending order).
Only the pairs with the highest scores are selected.
The user can choose the number of top combinations to be selected.

After the selection of a pair of rankers, which is denoted by $\mathfrak{X}^{*}_{2}$,  combinations of other sizes are selected by performing intersection and union operations, in a procedure detailed described in~\cite{paperUSRAF}.

For computing the correlation between rankers, the Rank-Biased Overlap (RBO)~\cite{PaperSimMeasureRank_Australia2010} measure is used.
This measure considers the overlap between top-$k$ lists at increasing depths.
The weight of the overlap is calculated based on probabilities defined at each depth.
It can be formally defined as follows:
%
\begin{equation}
\label{lblEqRBO}
	  \lambda(\tau_i,\tau_j,k,\alpha) =  (1-\alpha)  \sum_{d=1}^{k} \alpha^{d-1}  \times \frac{| \mathcal{N}(i,k) \cap \mathcal{N}(j,k) |}{d},
\end{equation}

\noindent where $\mathcal{N}(i,k)$ denotes the natural neighborhood of the top-$k$ images for $x_{i}$ and $\alpha$ is a constant ($\alpha = 0.9$ was used for all the experiments).

In this work, rather than the effectiveness estimations and the fusion method employed in~\cite{paperUSRAF}, we used a query performance prediction measure (HQPP) and a recent manifold ranking aggregation method (LHRR) that model the ranked lists through hypergraph structures~\cite{paperLHRR}.
Both are discussed in the next sub-sections.

\subsection{Hypergraph Query Performance Prediction}
\label{secHQPP}

Query Performance Prediction (QPP) can be broadly defined as the task of estimating the effectiveness of a search/retrieval operation performed in response to a query, where no labeled data is available~\cite{PaperQPP_TOIS16}. 
Initially proposed for textual retrieval systems~\cite{RankingRobust_CIKM2006,QPPWeb_SIGIR2007}, the task assumed a diversified taxonomy in the literature and has been establishing as a promising approach in image retrieval systems~\cite{PaperRelevPred_JVCIR19}.

In this work, we propose to use a Hypergraph Query Performance Prediction (HQPP) score for predicting the effectiveness of rankings produced by Person Re-ID features. The HQPP score uses a hypergraph formulation recently proposed~\cite{paperLHRR} for manifold ranking on multimedia retrieval.
Hypergraphs are a robust generalization of graphs, providing a powerful tool for capturing high-order relationships in several domains~\cite{PaperHyperRankSSL_CVPR2010,PaperSpectralHyper_NIPS2007,PaperSpecLearnHyper_KDD2008}.
In opposition to traditional graph-based approaches, which represent only pairwise relationships, hypergraphs allow connecting any number of nodes in order to represent similarity among sets of objects~\cite{BookBretto_Hypergraphs2013}. 

This work and the HQPP score use a hypergraph model mainly based on the following main hypotheses and ideas:
\begin{itemize}
    \item Similar objects present similar ranked lists and, therefore, similar hyperedges. Once the hyperedges are represented by an incidence matrix, the product of the hyperedges can be exploited to compute a more effective similarity measure between nodes; 
    \item Similar objects are expected to reference each other in the same hyperedge. Therefore, hyperedges that concentrate a high number of ranking references on a few nodes are expected to be more effective. The Hypergraph Query Performance Prediction (HQPP) is formally defined based on this  conjecture.
\end{itemize}

Following the definition of~\cite{paperLHRR}, a hypergraph can be defined as a tuple $H_G = (V, E, h_p)$, where $V$ represents a set of vertices and $E$ denotes the hyperedge set.
The set of hyperedges $E$ can be defined as a family of subsets of $V$ such that $\bigcup_{e \in E} = V$. 
To each hyperedge $e_i$, a positive score $h_p(e_i)$  denotes the confidence of relationships among a set of vertices established by the hyperedge $e_i$.

While graphs are commonly represented by adjacency matrices, hypergraphs are often represented by incidence matrices.
The incidence of a hyperedge $e_i$ on a vertice $v_j$ is represented by an incidence matrix $\mathbf{H}$, defined as follows:

\begin{equation}
	h (e_i, v_j) = 	\left\{ \begin{array}{ll}
		r(e_i,v_j), & ~\textrm{if } v_j \in e_i, \\
		0, & \textrm{ otherwise,}\\
 	\end{array} \right. 
\end{equation}	

\noindent where $h (e_i, v_j)$ denotes the reliance of the vertex $v_j$ to belong to a hyperedge $e_i$ and $r(e_i,v_j)$ is a function with a codomain in the $\mathbb{R}^+$ that indicates the degree to which the vertex $v_j$ belongs to a hyperedge $e_i$.
A hyperedge $e_i$ is defined for each image $x_i \in \mathcal{C}$ based on the $k$-neighborhood set of $x_i$ and its respective neighbors.
In this context, the function $r(e_i,v_j)$ is defined based on the density of ranking references to $v_j$ in the ranking of $x_i$ and its neighbors. 
Formally, the function is defined as:

\begin{equation}
\label{eqNr}
\begin{split}
	r(e_i,v_j) =  \sum_{y \in \mathcal{N}(i,k) \wedge j \in \mathcal{N}(y,k) }     \eta_r(i,y) \times \eta_r(y,j),
\end{split}  						 
\end{equation}

\noindent where $\eta_r(i,x)$ is a function that assigns a weight of relevance to image $x$ according to its position in the ranked list $\tau_i$.
 The weight assigned to $x$ according to its position in the ranked list $\tau_i$ is defined as follows:

\begin{equation}
	\eta_r (i, x) = 1-\log_k \tau_i(x).
\end{equation}


The size of the hyperedges varies according to the number of co-occurrence of images.
A high diversity of elements may indicate a high degree of uncertainty and this information will be exploited for defining the weights of hyperedges.
The weight of a hyperedge $h_p(e_i)$ denotes the confidence of relationships established among vertices by the hyperedge.

In order to compute the weight $h_p(e_i)$, we use the Hypergraph Neighborhood Set $\mathcal{N}_h$, which contains the $k$ vertices with the greatest $h(e_i,\cdot)$ scores in the hyperedge $e_i$.
As such, the hyperedge weight $h_p(e_i)$ is defined as:

\begin{equation}
	h_p (e_i) =  \sum_{j \in \mathcal{N}_h(i,k) }  h(i,j).
\end{equation}	

A high-effective hyperedge is expected to present an elevated value of $h_p$, indicating a consistent co-occurrence of the same elements with high confidence of membership. 
Therefore, the hyperedge weight $h_p (e_i)$ is defined as the Hypergraph Query Performance Prediction (HQPP) for ranked list of image $x_i$, which is denoted by $\gamma$:

\begin{equation}
	\gamma(\tau_i) = h_p (e_i).
\end{equation}	

For a given ranker $R_{i}$, the $\gamma$ can be computed for all the ranked lists to obtain the value of $\gamma(R_{i})$ in Equation~(\ref{eq:selection_eq}).
We highlight that, HQPP was used to define $\gamma$ in this work, but our approach is flexible and capable of supporting other measures.

\subsection{Hypergraph Manifold Rank Aggregation}

Once the person Re-ID features are selected, we fuse the respective produced rankings through a recently proposed manifold learning algorithm~\cite{paperLHRR}.
The Log-based Hypergraph of Ranking References (LHRR)~\cite{paperLHRR}, briefly described in this section, captures the dataset manifold structure through a hypergraph-based similarity measure, which can be used to rank aggregation tasks.

LHRR~\cite{paperLHRR} exploits the hypergraph formulation discussed in the last section to represent high-order similarity relationships encoded in the dataset manifold. Subsequently, pairwise similarity scores are computed, allowing more effective ranking results.
The pairwise similarity is computed based on the conjecture that  similar elements present similar hyperedge representations. The similarity between hyperedges is computed based on the product of the incidence matrix $\mathbf{H}$ and its transpose to encode reciprocal relationship. 
A pairwise similarity matrix $\mathbf{S}$ is computed as:

\begin{equation}
\label{eqHadamardSS}
\mathbf{S} = (\mathbf{H} \mathbf{H}^T )  \circ (\mathbf{H}^T \mathbf{H})
\end{equation}

In addition to the product of hyperedges, a Cartesian product operation is conducted to extract useful pairwise relationships direct from the set of elements defined by the hyperedges.
Given two hyperedges $e_q, e_i \in E$, the Cartesian product between them can be defined as:

\begin{equation}
 e_q \times e_i = \{ (v_x,v_y): ~ v_x \in e_q   \wedge   v_y \in e_i \}.
\end{equation}

Let ${e_q}^2$ denote the Cartesian product between the elements of the same hyperedge $e_q$, for each pair of vertices $(v_i,v_j) \in {e_q}^2$ a pairwise similarity relationship  $cp$ is computed to define the membership degrees of $v_i$ and $v_j$. The function is formally defined as:

\begin{equation}
cp (e_q,v_i,v_j) = h_p(e_q) \times h(e_q,v_i) \times h(e_q,v_j).
\end{equation}

A similarity measure based on Cartesian product is defined through a matrix $\mathbf{C}$, with each position computed as follows:

\begin{equation}
c(i,j) = \sum_{e_q \in E \wedge  (v_i,v_j) \in {e_q}^2 }  \;   cp (v_i,v_j).
\end{equation}

The pairwise similarity defined based on hyperedges and Cartesian product operations provides complementary information. Hence, an affinity matrix $\mathbf{W}$ is computed by combining both matrices as:

\begin{equation}
\label{eqHadamardW}
\mathbf{W} = \mathbf{C} \circ \mathbf{S}.
\end{equation}

Based on the affinity measure defined by $\mathbf{W}$, a ranking procedure can be performed for each feature giving rise to a new set of ranked lists.
Next, a multiplicative rank-based formulation is used to combine the features, exploiting an adaptive weight, which is assigned to each query/feature according to the weight of  the respective hyperedge. Let $\eta_f$ denote the fused affinity measure; each element is computed as follows considering the top-$L$ positions of $\tau_q$:

\begin{equation}
\eta_f(q,i) = \prod_{f=1}^{m} \frac{ (1+h_p(f,e_q)) } { (1+\log_L \tau_{q,f}(i)) },
\end{equation}

\noindent where $h_p(f,e_q)$ is the weight of hyperedge $e_q$ according to the feature $f$ and $\tau_{q,f}(i)$ denote the position of $x_i$ in the ranked list of $x_q$ according to the feature $f$.
The combined affinity measure $\eta_f(\cdot,\cdot)$ gives rise to a unique set of ranked lists which is re-processed by the LHRR~\cite{paperLHRR} algorithm as a single feature.

\section{Experimental Evaluation}
\label{secExpEval}

This section presents the experimental results conducted for evaluating our proposed approach.
The experimental evaluation was conducted on 4 datasets often used for person Re-ID with sizes ranging from 14,097 to 39,902.
For each dataset, up to 28 rankers were considered of different modalities (e.g. traditional descriptors, bag of visual words, deep learning).
The large number of rankers is used with the objective of evaluating the capacity of our selection approach.
It is desirable that, if the selection is accurate, only the most effective are selected to be fused.
Also, there is a very large number of possible combinations, when all the possible sizes are considered.
With 28 rankers, there are 268,435,456 possible combinations.
Since it is impractical to execute all of them, selection is of fundamental importance in this context.
The experimental evaluation also considers a comparison with fusion baselines and with state-of-the-art Re-ID approaches.

\subsection{Experimental Protocol}

Our experimental protocol considers 4 datasets, which are detailed in Table~\ref{tab:datasets}.
For CUHK03~\cite{cuhk03}, we considered the detected version of the dataset, which uses the bounding boxes extracted by a DPM detector, and adopted the experimental protocol proposed by~\cite{cuhk03_new_protocol}.
Both Market1501~\cite{market1501} and DukeMTMC~\cite{dukemtmc} use the experimental protocol proposed by the original authors that published the datasets.
In the case of the Airport~\cite{airport}, a different protocol was adopted, where all the images (training, probe, and gallery) are considered as queries (probes) and part of the gallery at the same time.
For all of them, the MAP (Mean Average Precision) was reported.
The R1 corresponds to the first value of the CMC (Cumulative Matching Characteristics) curve, which indicates the number of ranked lists that have an image that corresponds to the same individual in the first position after the query image.
All the evaluations are single-shot (single-query), this means that each query corresponds to only one image.

\begin{table}[ht!]
\caption{Datasets considered in the experimental evaluation.}
\label{tab:datasets}
\centering
\resizebox{.84\textwidth}{!}{ 
\begin{tabular}{|l|c|c|c|c|c|}
\hline
\textbf{Dataset}    & \textbf{People} & \textbf{BBox} & \textbf{Cam} & \textbf{Label} & \textbf{Evaluation} \\ \hline
\textbf{CUHK03~\cite{cuhk03, cuhk03_new_protocol}}     & 1,467           & 14,097        & 2            & DPM            & R1/ MAP~\cite{cuhk03_new_protocol}          \\ \hline
\textbf{Market1501~\cite{market1501}} & 1,501           & 32,217        & 6            & DPM            & R1/MAP~\cite{market1501}          \\ \hline
\textbf{DukeMTMC~\cite{dukemtmc}}   & 1,812           & 36,411        & 8            & Manual         & R1/MAP~\cite{dukemtmc}          \\ \hline
\textbf{Airport~\cite{airport}}    & 9,651           & 39,902        & 6            & ACF            & MAP (all queries) \\ \hline
\end{tabular}
}
\end{table}

Table~\ref{tab:descriptors} presents all the descriptors used in the experimental evaluation (i.e., as the input of our method).
The number of rankers per dataset ranges from 21 to 28, and emcompasses different modalities (e.g. statistical, bag of visual words, deep learning).
For most of the extractions, the Euclidean distance was considered, no metric learning was applied aiming at keeping the protocol fully unsupervised (not using the labels from the dataset being evaluated).
Except for OSNET-AIN, where cosine distance was used, as done in~\cite{paperOSNET-IBN-AIN}.
For the non-deep methods (GBICOV, LOMO, GOG, WHOS, ELF, HLBP, SDC, and BOVW), we applied PCA to reduce the feature vectors to 100 dimensions before the distances were extracted, as also done by other authors~\cite{airport}.
In total, we considered 7 different Convolutional Neural Networks (CNN): MobileNetV2, RESNET50, HACNN, MLFN, OSNET, OSNET-IBN, and OSNET-AIN.
For each of these models, three different trainings were performed:
on Market dataset, which is indicated by (M); 
on DukeMTMC dataset, which is indicated by (D); 
and on MSMT17 dataset, which is indicated by (MT).
Different trainings and models were used with the objective of evaluating our selection and fusion.
Most of them were pre-trained on the ImageNet~\cite{paperImagenet} dataset, except HACNN which was trained on Re-ID datasets from scratch.
The MSMT17~\cite{paperMSMT17} was used to train some of the networks, since it is a large dataset (126,441 images of 4,101 people in 15 cameras) that can facilitate the capacity of generalization.
The majority of the networks were trained considering only the training subset, but RESNET50 and OSNET variants used all of the images (train, probe, and gallery) when trained on MSMT17.
The CNN features were extracted with the Torchreid~\cite{paperTorchreid} trained weights~\footnote{\url{https://kaiyangzhou.github.io/deep-person-reid/MODEL_ZOO.html}}.

To keep the comparisons fair, \textbf{since our evaluation is unsupervised, the models that were trained on the target dataset were not used.}
For example, train on Market and test on Market is removed.
These cases are reported with ``---''.

\begin{table}[ht!]
\centering
\caption{Values of MAP (\%) and R-01 (\%) for the descriptors and datasets considered on the experiments. The dataset used to train the CNN is referenced between parentheses (M = Market, D = DukeMTMC, MT = MSMT17).}
\label{tab:descriptors}
\resizebox{.83\textwidth}{!}{ 
\begin{tabular}{l|c|c||c|c||c|c||c|}
\cline{2-8}
\textbf{}                                       & \multicolumn{7}{c|}{\textbf{Datasets}}                                                                                                                                                                                                                     \\ \cline{2-8} 
\multicolumn{1}{c|}{}                           & \multicolumn{2}{c||}{\textbf{CUHK03}}                                 & \multicolumn{2}{c||}{\textbf{Market1501}}                             & \multicolumn{2}{c||}{\textbf{DukeMTMC}}                               & \multicolumn{1}{c|}{\textbf{Airport}} \\ \hline
\multicolumn{1}{|l|}{\textbf{Features}}         & \multicolumn{1}{c|}{\textbf{R1}} & \multicolumn{1}{c||}{\textbf{MAP}} & \multicolumn{1}{c|}{\textbf{R1}} & \multicolumn{1}{c||}{\textbf{MAP}} & \multicolumn{1}{c|}{\textbf{R1}} & \multicolumn{1}{c||}{\textbf{MAP}} & \multicolumn{1}{c|}{\textbf{MAP}}     \\ \hline
\multicolumn{1}{|l|}{\textbf{GBICOV~\cite{paperGBICOV}}}           & 0.63                             & 0.82                              & 10.21                            & 3.27                              & ---                              & ---                               & ---                                   \\ \hline
\multicolumn{1}{|l|}{\textbf{LOMO~\cite{paperLOMO}}}             & 0.79                             & 0.89                              & 19.15                            & 6.46                              & 6.60                             & 2.82                              & 35.35                                 \\ \hline
\multicolumn{1}{|l|}{\textbf{GOG~\cite{paperGOG}}}              & 0.49                             & 0.77                              & 21.56                            & 7.55                              & 10.82                            & 4.40                              & 34.11                                 \\ \hline
\multicolumn{1}{|l|}{\textbf{WHOS~\cite{paperWHOS}}}             & 0.39                             & 0.56                              & 20.01                            & 6.23                              & 7.50                             & 2.65                              & 34.75                                 \\ \hline
\multicolumn{1}{|l|}{\textbf{ELF~\cite{paperELF}}}              & 0.34                             & 0.52                              & 12.02                            & 3.85                              & 2.42                             & 0.83                              & 31.17                                 \\ \hline
\multicolumn{1}{|l|}{\textbf{HLBP~\cite{paperHLBP}}}             & 0.32                             & 0.43                              & 7.07                             & 2.18                              & 0.76                             & 0.54                              & 32.68                                 \\ \hline
\multicolumn{1}{|l|}{\textbf{SDC~\cite{paperSDC}}}              & 0.18                             & 0.34                              & 11.02                            & 3.78                              & 2.96                             & 1.18                              & 31.57                                 \\ \hline
\multicolumn{1}{|l|}{\textbf{BOVW-350~\cite{market1501}}}         & 1.69                             & 1.80                              & 33.11                            & 13.34                             & 14.41                            & 6.71                              & 32.73                                 \\ \hline
\multicolumn{1}{|l|}{\textbf{BOVW-500~\cite{market1501}}}         & 1.56                             & 1.81                              & 32.33                            & 12.94                             & 14.14                            & 6.68                              & 33.09                                 \\ \hline
\multicolumn{1}{|l|}{\textbf{MobileNetV2 (M)~\cite{paperMobileNetv2}}}  & 4.39                             & 4.34                              & ---                              & ---                               & 24.01                            & 12.34                             & 35.62                                 \\ \hline
\multicolumn{1}{|l|}{\textbf{MobileNetV2 (D)~\cite{paperMobileNetv2}}}  & 4.30                             & 4.30                              & 37.80                            & 15.63                             & ---                              & ---                               & 37.23                                 \\ \hline
\multicolumn{1}{|l|}{\textbf{MobileNetV2 (MT)~\cite{paperMobileNetv2}}} & 8.87                             & 8.51                              & 37.86                            & 16.56                             & 42.59                            & 23.79                             & 38.84                                 \\ \hline
\multicolumn{1}{|l|}{\textbf{RESNET50 (M)~\cite{paperRESNET}}}     & 3.84                             & 3.90                              & ---                              & ---                               & 25.67                            & 13.62                             & 38.01                                 \\ \hline
\multicolumn{1}{|l|}{\textbf{RESNET50 (D)~\cite{paperRESNET}}}     & 5.84                             & 5.85                              & 42.64                            & 18.39                             & ---                              & ---                               & 40.25                                 \\ \hline
\multicolumn{1}{|l|}{\textbf{RESNET50 (MT)~\cite{paperRESNET}}}    & 13.68                            & 13.08                             & 46.59                            & 22.82                             & 52.29                            & 32.00                             & 41.95                                 \\ \hline
\multicolumn{1}{|l|}{\textbf{HACNN (M)~\cite{paperHACNN}}}        & 5.51                             & 5.69                              & ---                              & ---                               & 23.79                            & 13.13                             & 36.40                                 \\ \hline
\multicolumn{1}{|l|}{\textbf{HACNN (D)~\cite{paperHACNN}}}        & 3.11                             & 3.28                              & 43.74                            & 18.87                             & ---                              & ---                               & 38.85                                 \\ \hline
\multicolumn{1}{|l|}{\textbf{HACNN (MT)~\cite{paperHACNN}}}       & 9.71                             & 9.68                              & 49.23                            & 23.30                             & 42.19                            & 25.57                             & 42.94                                 \\ \hline
\multicolumn{1}{|l|}{\textbf{MLFN (M)~\cite{paperMLFN}}}         & 4.91                             & 5.19                              & ---                              & ---                               & 30.39                            & 16.96                             & 38.67                                 \\ \hline
\multicolumn{1}{|l|}{\textbf{MLFN (D)~\cite{paperMLFN}}}         & 4.72                             & 4.74                              & 45.55                            & 20.26                             & ---                              & ---                               & 40.15                                 \\ \hline
\multicolumn{1}{|l|}{\textbf{MLFN (MT)~\cite{paperMLFN}}}        & 10.58                            & 10.19                             & 46.59                            & 21.98                             & 48.70                            & 28.98                             & 41.17                                 \\ \hline
\multicolumn{1}{|l|}{\textbf{OSNET (MT)~\cite{paperOSNET}}}       & 20.83                            & 19.84                             & 65.94                            & 37.36                             & 65.98                            & 45.20                             & 45.47                                 \\ \hline
\multicolumn{1}{|l|}{\textbf{OSNET-IBN (M)~\cite{paperOSNET-IBN-AIN}}}    & 10.48                            & 10.22                             & ---                              & ---                               & 48.52                            & 26.59                             & 40.96                                 \\ \hline
\multicolumn{1}{|l|}{\textbf{OSNET-IBN (D)~\cite{paperOSNET-IBN-AIN}}}    & 8.01                             & 7.85                              & 57.48                            & 26.01                             & ---                              & ---                               & 40.65                                 \\ \hline
\multicolumn{1}{|l|}{\textbf{OSNET-IBN (MT)~\cite{paperOSNET-IBN-AIN}}}   & 21.70                            & 20.78                             & 66.45                            & 37.13                             & 67.41                            & 45.52                             & 45.37                                 \\ \hline
\multicolumn{1}{|l|}{\textbf{OSNET-AIN (M)~\cite{paperOSNET-IBN-AIN}}}    & 12.14                            & 11.67                             & ---                              & ---                               & 52.42                            & 30.35                             & 42.05                                 \\ \hline
\multicolumn{1}{|l|}{\textbf{OSNET-AIN (D)~\cite{paperOSNET-IBN-AIN}}}    & 9.54                             & 9.24                              & 61.10                            & 30.64                             & ---                              & ---                               & 42.79                                 \\ \hline
\multicolumn{1}{|l|}{\textbf{OSNET-AIN (MT)~\cite{paperOSNET-IBN-AIN}}}   & 28.49                            & 27.00                             & 69.95                            & 43.30                             & 71.14                            & 52.69                             & 52.26                                 \\ \hline
\end{tabular}
}
\end{table}

\subsection{Experimental Analysis}

\begin{figure}[ht!]
    \centering
    \includegraphics[width=.55\textwidth]{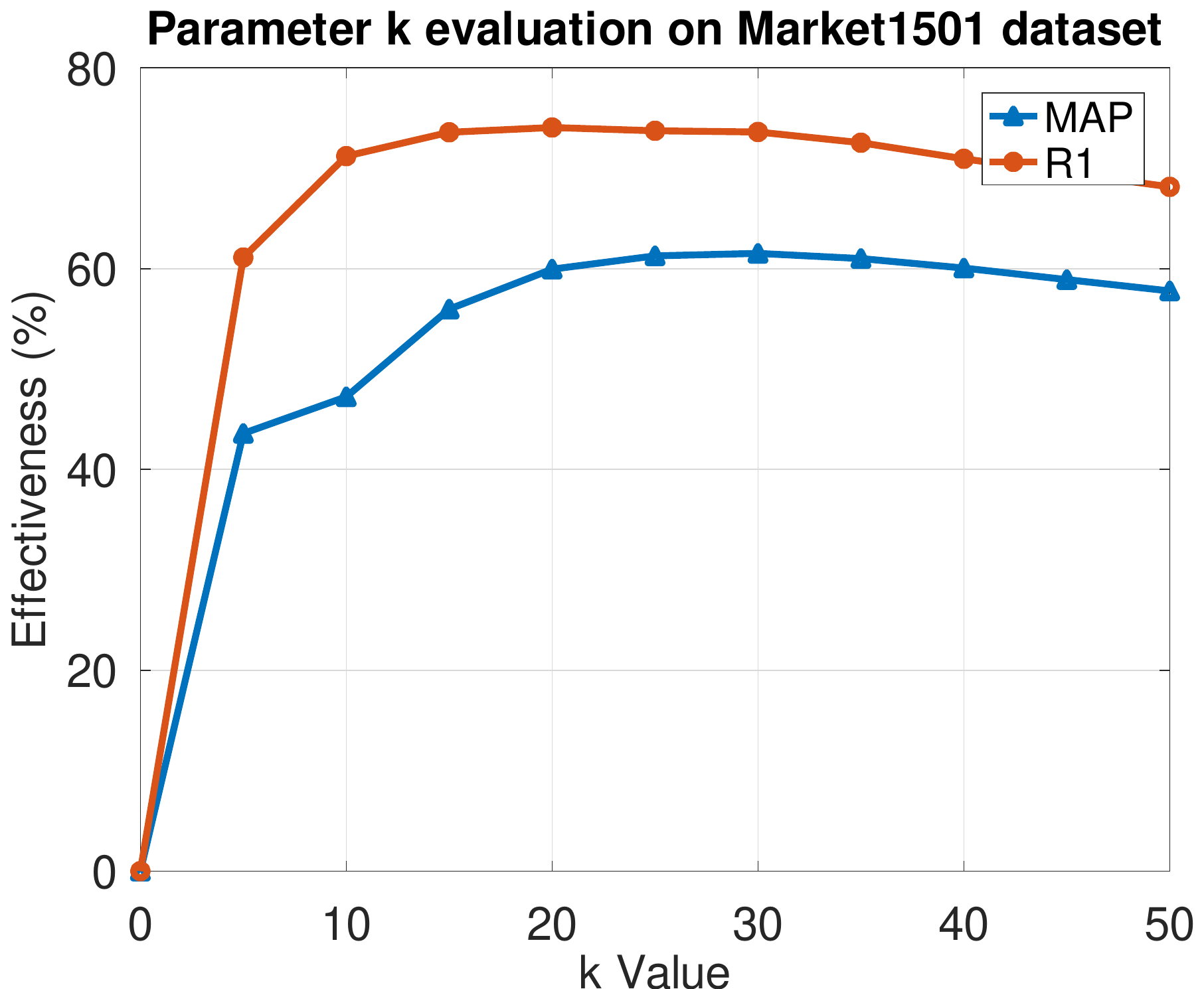}
    \caption{Evaluation of the impact of parameter $k$ on MAP and R1 for Market1501 dataset.}
    \label{fig:k_analysis}
\end{figure}

The neighborhood size, denoted by $k$, is used in multiple steps of our method:
for calculating the effectiveness measure,
for the correlation measure,
and in the fusion stage.
Figure~\ref{fig:k_analysis} presents an experiment that was conducted to evaluate the impact of $k$ on Market1501 dataset, where both R1 and MAP are shown for different values of $k$.
Notice that the method is robust to different parameter settings.
We used $k= 20$ for CUHK03, Market1501, DukeMTMC and $k = 10$ for Airport in all of the remaining experiments.
For the Airport dataset, a smaller $k$ seems to be more adequate, since it has less images per individual (around 4) compared to the other collections.

\begin{figure}[ht!]
    \centering
    \includegraphics[width=.55\textwidth]{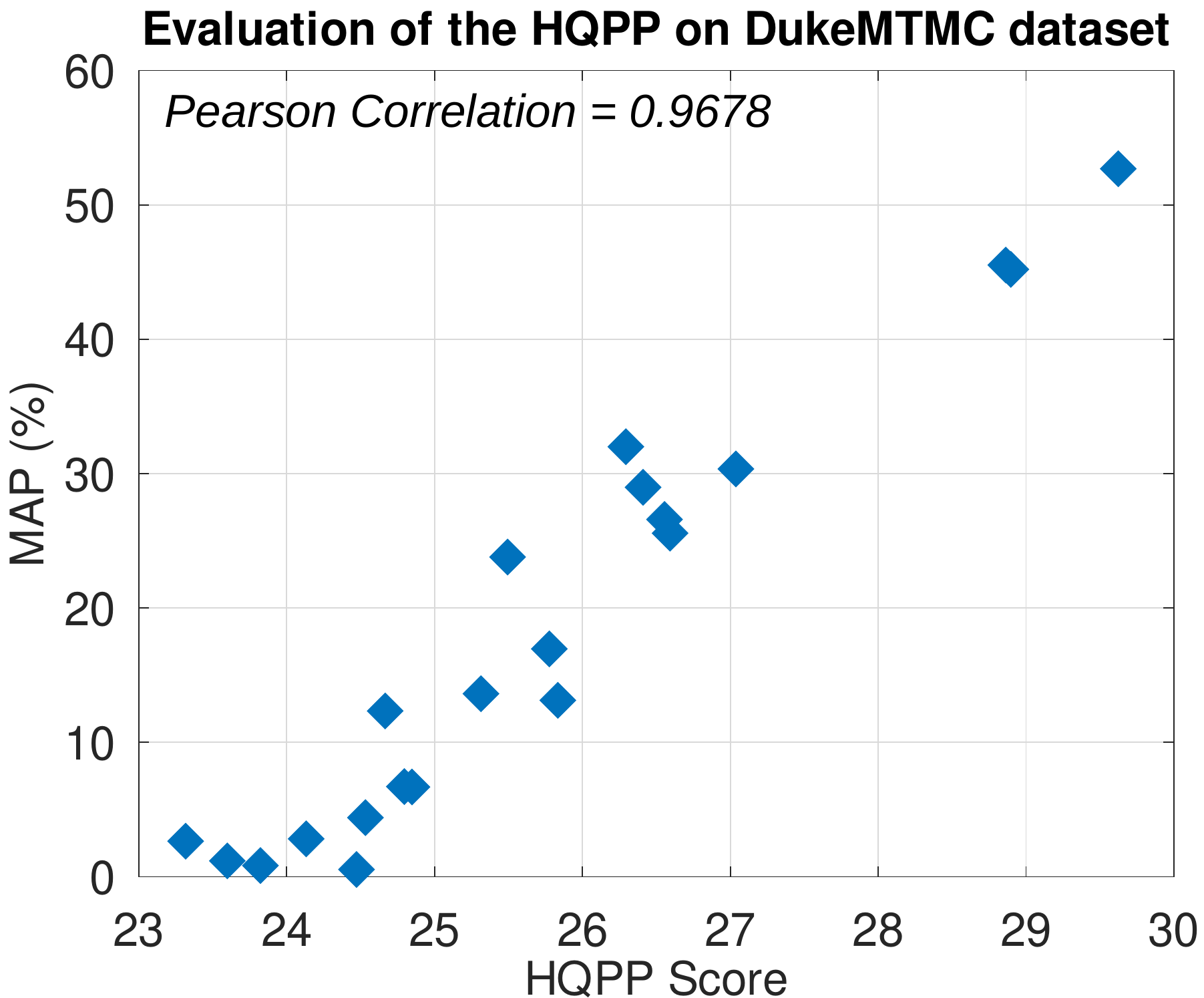}
    \caption{Evaluation of the HQPP measure compared to the MAP on DukeMTMC dataset.}
    \label{fig:hqpp_analysis}
\end{figure}

In this work, HQPP is proposed as a measure to estimate the quality of each ranker in the selection stage.
Aiming at assessing the use of this measure in Re-ID scenarios, Figure~\ref{fig:hqpp_analysis} shows an experiment where each dot corresponds to a different ranker and the MAP (measure that uses labeled data) is compared to the HQPP performance prediction score.
As can be seen in the graph, there is a high correlation between the measures.
The Pearson correlation among the dots is 0.9678, which indicates the high effectiveness of the selection strategy.

\begin{figure}[!ht]
    \centering
    \includegraphics[width=.6\textwidth]{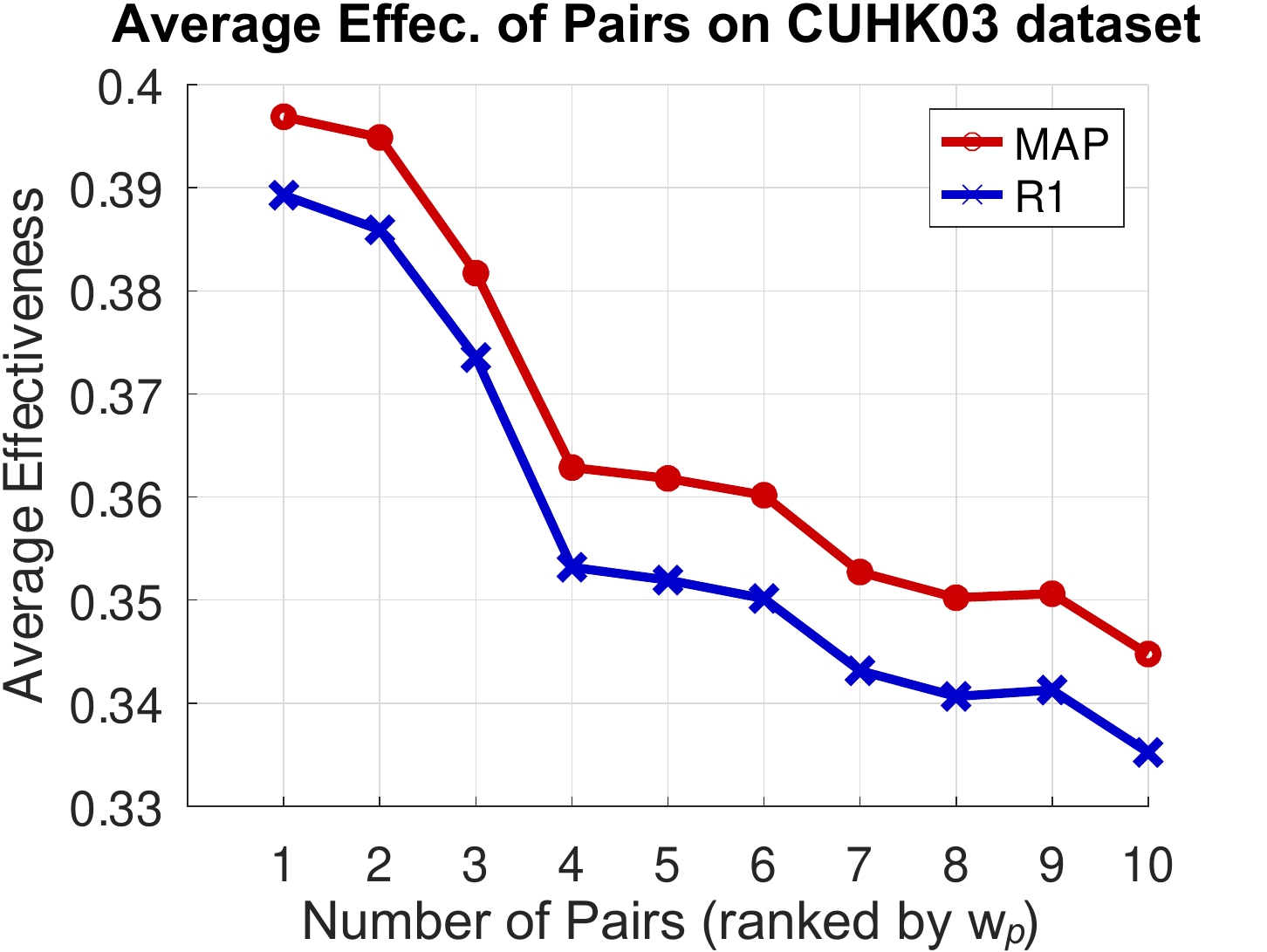}
    \caption{Average MAP of top pairs on CUHK03 dataset.}
    \label{fig:pairs_cuhk03}
\end{figure}

\begin{figure}[!ht]
    \centering
    \includegraphics[width=.6\textwidth]{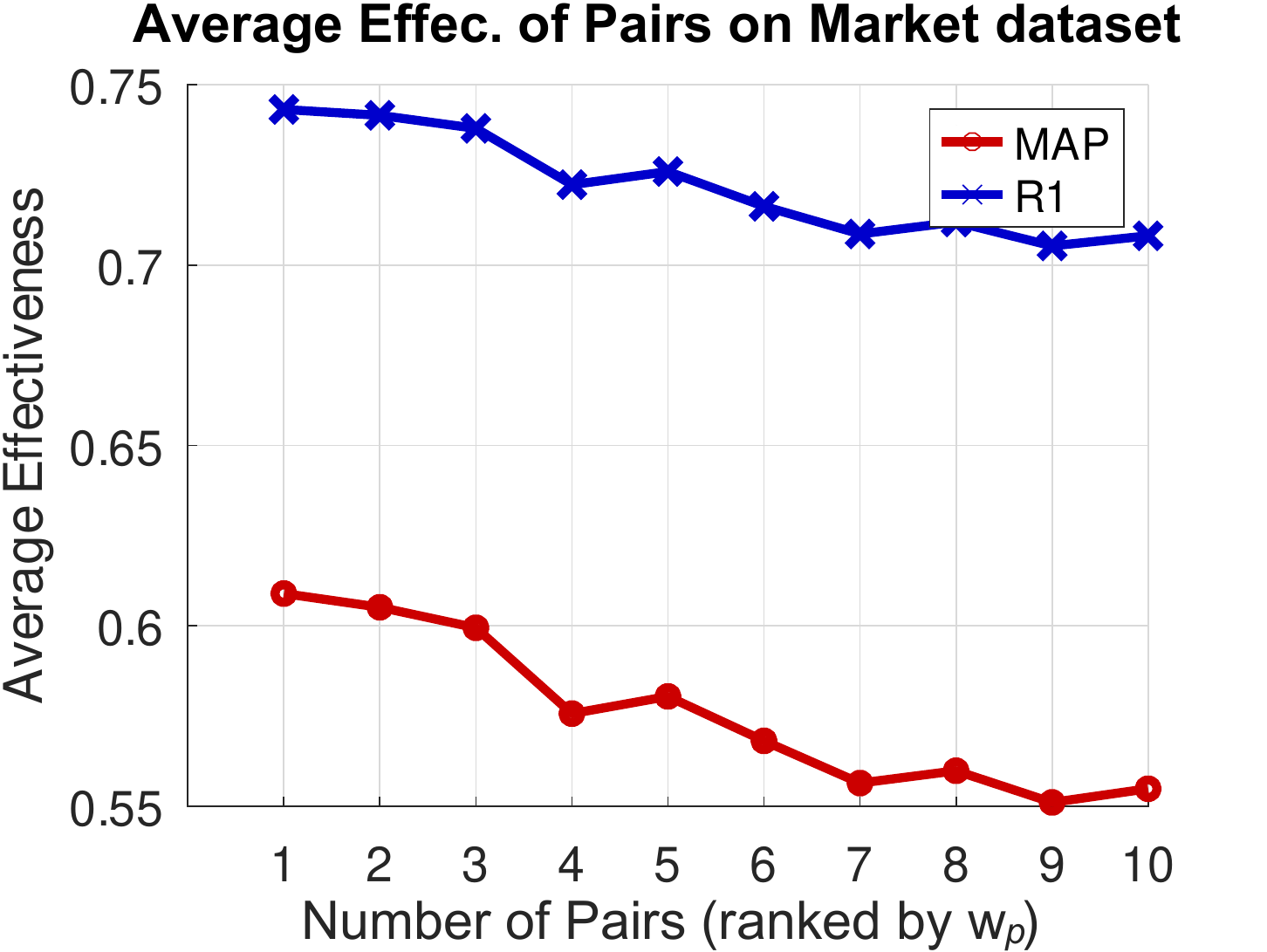}
    \caption{Average MAP of top pairs on Market dataset.}
    \label{fig:pairs_market}
\end{figure}

\begin{figure}[!ht]
    \centering
    \includegraphics[width=.6\textwidth]{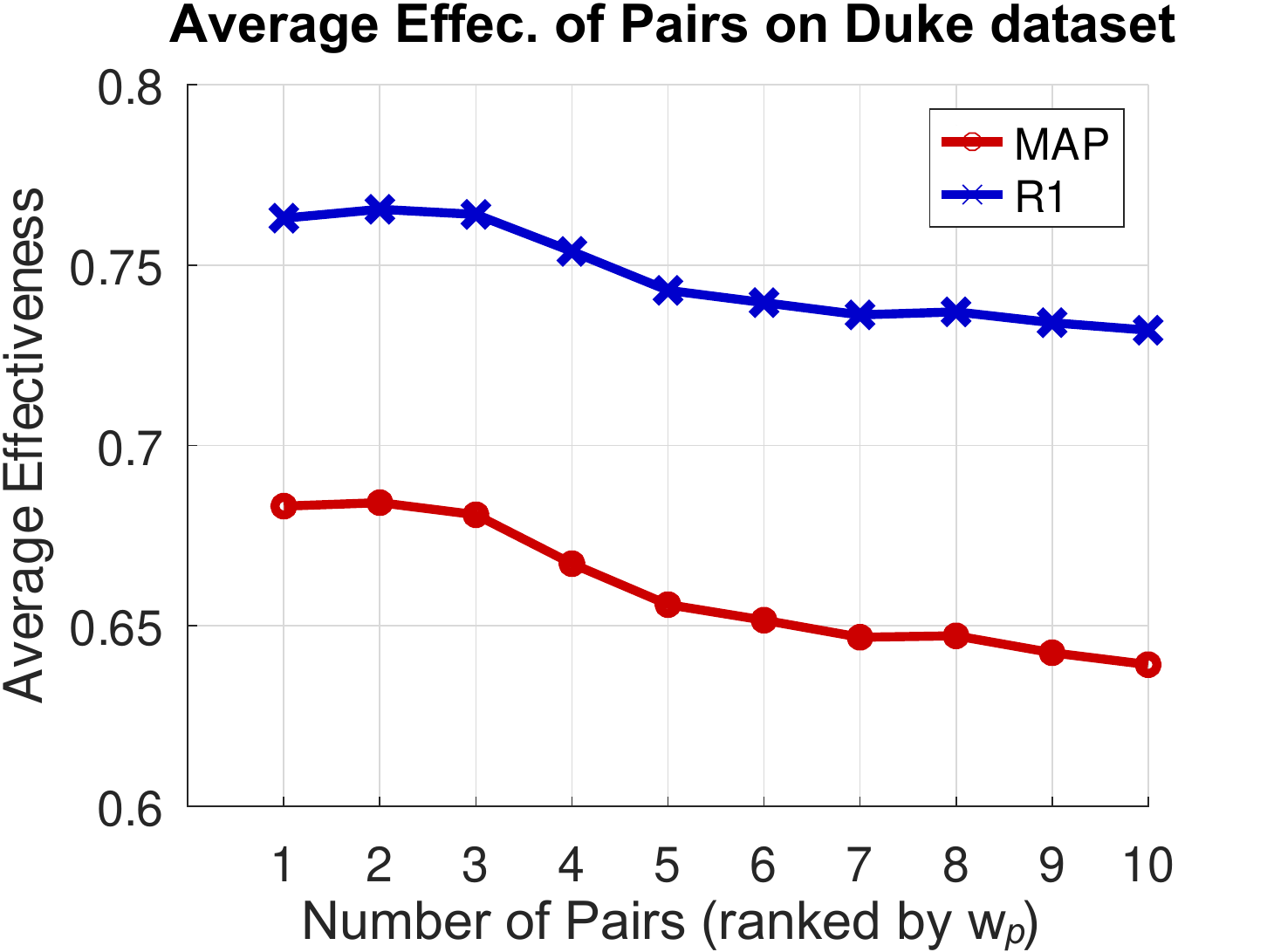}
    \caption{Average MAP of top pairs on Duke dataset.}
    \label{fig:pairs_duke}
\end{figure}

The HRSF ranks the best combinations for each size.
The user can choose the number of top combinations to be selected.
We conducted an experiment on CUHK03, Market, and DukeMTMC (Figures~\ref{fig:pairs_cuhk03},~\ref{fig:pairs_market}, and~\ref{fig:pairs_duke}) where the average MAP and R1 of the selected ranker pairs is presented as the number of selected pairs changes.
Notice that the highest MAP and R1 values are in the first position (top-1), which evinces that the combination with the highest $w_{p}$ (ranked in the first position) is also the one with highest effectiveness.

The best combination available in the top-5 for each size is reported in Table~\ref{tab:combinations}.
We report sizes from 1 to 6 ($\mathfrak{X}^{*}_{2}$, ..., $\mathfrak{X}^{*}_{6}$) and the best combination among them (which can be denoted just as $\mathfrak{X}^{*}$) is highlighted in bold. 
The best isolated ranker in each case is also listed for comparison purposes and to facilitate the visualization of the relative MAP gain.
Notice that OSNET, OSNET-AIN, and OSNET-IBN are the most commonly selected rankers, which evinces the effectiveness of our selection, once these rankers are among the most effective ones. 
Additionally, in all cases the proposed selection and fusion \textbf{achieves better results the the best ranker in isolation}.
The complementarity among the methods can be exploited by our approach achieving gains up to +47\% (MAP) after the selection and fusion is performed.
The results also indicate that gains can be obtained when networks trained on different datasets are combined, even when the same architecture is used.

\definecolor{lightgray}{rgb}{0.88, 0.88, 0.88}

\begin{table}[H]
\centering
\caption{The best selected combination of each size (among top-5) is reported on each dataset.}
\label{tab:combinations}
\resizebox{.99\textwidth}{!}{
\begin{tabular}{|c|c|c|c|c|c|}
\hline
\multirow{2}{*}{\textbf{Dataset}}                & \textbf{Comb.} & \multirow{2}{*}{\textbf{Selected and Fused Rankers}}                                                                                                     & \textbf{R1} & \multicolumn{1}{c|}{\textbf{MAP}} & \textbf{MAP} \\
               &  \textbf{Size}  &                                                                                                      & \multicolumn{1}{c|}{\textbf{(\%)}} & \multicolumn{1}{c|}{\textbf{(\%)}} & \textbf{R. Gain} \\ \hline
\hline
\multirow{9}{*}{\textbf{CUHK03}}                 & Best $R$                         & OSNET-AIN (MT)                                                                                                                          &  28.49  &  27.00 &   ---                        \\ \cline{2-6}
\multicolumn{1}{|l|}{\textbf{}} & $\mathfrak{X}^{*}_{2}$                             & \textbf{OSNET-AIN (MT) + OSNET-IBN (MT)}                                                                                                         &  \textbf{39.04}   &  \textbf{39.69}  & \textbf{+47.00\%}                        \\ \cline{2-6}
\multicolumn{1}{|l|}{\textbf{}} & $\mathfrak{X}^{*}_{3}$                          & \begin{tabular}[c]{@{}c@{}}OSNET-AIN (MT) + OSNET-IBN (MT) + OSNET (MT)\end{tabular}                                                  & 39.13         & 39.58  &  +46.59\%             \\ \cline{2-6}
\multicolumn{1}{|l|}{\textbf{}} & $\mathfrak{X}^{*}_{4}$                              & \begin{tabular}[c]{@{}c@{}}OSNET-AIN (MT) + OSNET-IBN (MT) + OSNET (MT)\\ + OSNET-AIN (M)\end{tabular}                                  &  38.02    &  38.80 &  +43.70\%                         \\ \cline{2-6}
\multicolumn{1}{|l|}{\textbf{}} & $\mathfrak{X}^{*}_{5}$                             & \begin{tabular}[c]{@{}c@{}}OSNET-AIN (MT) + OSNET-IBN (MT) + OSNET (MT)\\ + OSNET-AIN (M) + OSNET-IBN (M)\end{tabular}                &  36.15       & 37.11 &    +37.44\%                       \\ \cline{2-6}
\multicolumn{1}{|l|}{\textbf{}} & $\mathfrak{X}^{*}_{6}$                              & \begin{tabular}[c]{@{}c@{}}HACNN (MT) +  OSNET-AIN (D)  + OSNET-AIN (M)\\ + OSNET-AIN (MT) + OSNET-IBN (MT) + OSNET (MT)\end{tabular} & 35.46 &  36.17 &  +33.96\%                         \\ \hline
\hline
\multirow{9}{*}{\textbf{Market1501}}                 & Best $R$                          & OSNET-AIN (MT)                                                                                                                          &  69.95       &  43.30   &  ---                       \\ \cline{2-6}
\textbf{}                       & $\mathfrak{X}^{*}_{2}$                    & OSNET-AIN (MT) + OSNET (MT)                                                                                                             & 74.32        & 60.89    &  +40.62\%   \\ \cline{2-6}
\textbf{}                       & $\mathfrak{X}^{*}_{3}$                             & \begin{tabular}[c]{@{}c@{}}OSNET-AIN (MT) + OSNET (MT) + OSNET-AIN (D)\end{tabular}                                                   &  75.56      &  62.64                  &  +44.67\%  \\ \cline{2-6}
\textbf{}                       & $\mathfrak{X}^{*}_{4}$                             & \textbf{\begin{tabular}[c]{@{}c@{}}OSNET-AIN (MT) + OSNET (MT) + OSNET-AIN (D)\\ + OSNET-IBN (MT)\end{tabular}}                                 &  \textbf{75.71}        &   \textbf{62.94}                         &  \textbf{+45.36\%}  \\ \cline{2-6}
\textbf{}                       & $\mathfrak{X}^{*}_{5}$                            & \begin{tabular}[c]{@{}c@{}}OSNET-AIN (MT) + OSNET (MT) + OSNET-AIN (D)\\ + OSNET-IBN (MT) + HACNN (D)\end{tabular}                    &  74.00         &   60.69                         &  +40.16\%  \\ \cline{2-6}
\textbf{}                       & $\mathfrak{X}^{*}_{6}$                              & \begin{tabular}[c]{@{}c@{}}HACNN (MT) + OSNET-AIN (D) + OSNET-AIN (MT)\\ +  OSNET-IBN (D) + OSNET-IBN (MT) + OSNET (MT)\end{tabular}  &  73.57         & 59.85 &     +38.22\%                       \\ \hline
\hline
\multirow{9}{*}{\textbf{DukeMTMC}}                   & Best $R$                          & OSNET-AIN (MT)                                                                                                                          &   71.14     &  52.69                            & --- \\ \cline{2-6}
\textbf{}                       & $\mathfrak{X}^{*}_{2}$               & OSNET-AIN (MT) + OSNET-IBN (MT)                                                                                                         & 76.80        &    68.51                        &  +30.02\% \\ \cline{2-6}
\textbf{}                       & $\mathfrak{X}^{*}_{3}$                     & \textbf{\begin{tabular}[c]{@{}c@{}}OSNET-AIN (MT)  + OSNET-IBN (MT) + OSNET (MT)\end{tabular}}                                                 &  \textbf{77.24}        &  \textbf{68.88}                          &  \textbf{+30.73\%}  \\ \cline{2-6}
\textbf{}                       & $\mathfrak{X}^{*}_{4}$                 & \begin{tabular}[c]{@{}c@{}}OSNET-AIN (MT)  + OSNET-IBN (MT) + OSNET (MT)\\ + RESNET (MT)\end{tabular}                                   &  76.89       &    68.56                         &  +30.12\%  \\ \cline{2-6}
\textbf{}                       & $\mathfrak{X}^{*}_{5}$                 & \begin{tabular}[c]{@{}c@{}}OSNET-AIN (MT)  + OSNET-IBN (MT) + OSNET (MT)\\ + RESNET (MT) + OSNET-AIN (M)\end{tabular}                 & 76.39          & 67.72                           &  +28.53\%  \\ \cline{2-6}
\textbf{}                       & $\mathfrak{X}^{*}_{6}$                 & \begin{tabular}[c]{@{}c@{}}OSNET-AIN (MT)  + OSNET-IBN (MT) + OSNET (MT)\\ + RESNET (MT) + OSNET-AIN (M) + MLFN (MT)\end{tabular}     & 75.90        &  66.96          &   +27.08\%                 \\ \hline
\hline
\multirow{9}{*}{\textbf{Airport}}                & Best $R$                          & OSNET-AIN (MT)                                                                                                                          &    ---         &  52.26         &  --- \\ \cline{2-6}
\textbf{}                       & $\mathfrak{X}^{*}_{2}$                             & OSNET-AIN (MT) + OSNET (MT)                                                                                                             &    ---     & 52.43         &  +0.33\%   \\ \cline{2-6}
\textbf{}                       & $\mathfrak{X}^{*}_{3}$                           & \begin{tabular}[c]{@{}c@{}}OSNET-AIN (MT) + OSNET (MT) + OSNET-IBN (MT)\end{tabular}                                                  &   ---      & 53.38         &  +2.14\%  \\ \cline{2-6}
\textbf{}                       & $\mathfrak{X}^{*}_{4}$                            & \begin{tabular}[c]{@{}c@{}}OSNET-AIN (MT) + OSNET (MT) + OSNET-IBN (MT)\\ + OSNET-AIN (M)\end{tabular}                                  & ---        & 53.91         &  +3.16\%  \\ \cline{2-6}
\textbf{}                       & $\mathfrak{X}^{*}_{5}$                            & \textbf{\begin{tabular}[c]{@{}c@{}}OSNET-AIN (MT) + OSNET (MT) + OSNET-IBN (MT)\\ + OSNET-AIN (M) + HACNN (MT)\end{tabular}}                   &   ---    & \textbf{54.09}       &  \textbf{+3.50\%}  \\ \cline{2-6}
\textbf{}                       & $\mathfrak{X}^{*}_{6}$                             & \begin{tabular}[c]{@{}c@{}}OSNET-AIN (MT) + OSNET (MT) + OSNET-IBN (MT)\\ + OSNET-AIN (M) + HACNN (MT) + MLFN (MT)\end{tabular}       &   ---      & 54.02  &   +3.37\%     \\ \hline
\end{tabular}}
\end{table}

\subsection{Comparison with Fusion Baselines}

In order to evaluate the proposed method compared to other approaches that both select and fuse the input features,
Table~\ref{tab:fusion_baselines} presents the proposed approach, HRSF, compared to both early and late fusion baselines on the four datasets.
In all the cases, the same set of features, which were presented in the experimental protocol, were used.
For the early fusion methods, the default parameters were used and all the features were processed with PCA to reduce the feature vectors to 100 components.
From all the features, the top-1000 were selected to compose the new feature vector and the Euclidean distance was computed.
As can be seen, the results of our approach are superior in most cases (CUHK03, Market1501, Airport) and comparable in others (DukeMTMC).

\begin{table}[H]
\centering
\caption{Proposed approach compared to early and late fusion baselines.}
\label{tab:fusion_baselines}
\resizebox{.76\textwidth}{!}{
\begin{tabular}{|c|c|c|c|c|}
\hline
\textbf{Dataset} & \textbf{Category} & \textbf{Method} & \textbf{R1 (\%)} & \textbf{MAP (\%)} \\
\hline
\hline
\multirow{4}{*}{\textbf{CUHK03}} & \multirow{2}{*}{\textbf{Early Fusion}}  & Laplace~\cite{paperLaplaceScore_2005}                                                                                                &  9.56   &  10.19                       \\

\multicolumn{1}{|l|}{\textbf{}} & & SPEC~\cite{paperSPEC_2007} &  9.29  & 9.97                       \\ 
\cline{2-5}
\multicolumn{1}{|l|}{\textbf{}} & \multirow{2}{*}{\textbf{Late Fusion}} & USRF~\cite{paperUSRAF} &  38.24 & 39.03                       \\ 
\multicolumn{1}{|l|}{\textbf{}} & & \cellcolor{lightgray} \textbf{HRSF (ours)} &  \cellcolor{lightgray} \textbf{39.04}       & \cellcolor{lightgray} \textbf{39.69}                      \\
\hline
\hline
\multirow{4}{*}{\textbf{Market1501}}  & \multirow{2}{*}{\textbf{Early Fusion}} & Laplace~\cite{paperLaplaceScore_2005}                                                                                                &  \textbf{82.07}   &  61.26                        \\ 

\multicolumn{1}{|l|}{\textbf{}} & & SPEC~\cite{paperSPEC_2007} & 77.14 & 54.90                    \\
\cline{2-5}
\multicolumn{1}{|l|}{\textbf{}} & \multirow{2}{*}{\textbf{Late Fusion}} & USRF~\cite{paperUSRAF} & 75.97 & 62.69                      \\
\multicolumn{1}{|l|}{\textbf{}} & & \cellcolor{lightgray} \textbf{HRSF (ours)} &  \cellcolor{lightgray} 75.71       & \cellcolor{lightgray} \textbf{62.94}                      \\
\hline
\hline
\multirow{4}{*}{\textbf{DukeMTMC}}  & \multirow{2}{*}{\textbf{Early Fusion}} & Laplace~\cite{paperLaplaceScore_2005}                                                                                                &  59.29   &  43.56                          \\

\multicolumn{1}{|l|}{\textbf{}} & & SPEC~\cite{paperSPEC_2007} &  59.29   &  43.56                      \\ 
\cline{2-5}
\multicolumn{1}{|l|}{\textbf{}} & \multirow{2}{*}{\textbf{Late Fusion}} & USRF~\cite{paperUSRAF} &  \textbf{77.82} &  \textbf{68.98}                     \\
\multicolumn{1}{|l|}{\textbf{}} & & \cellcolor{lightgray} \textbf{HRSF (ours)} &  \cellcolor{lightgray} 77.24       &  \cellcolor{lightgray} 68.88                      \\
\hline
\hline
\multirow{4}{*}{\textbf{Airport}}  & \multirow{2}{*}{\textbf{Early Fusion}} & Laplace~\cite{paperLaplaceScore_2005}                                                                                                &  ---   &  45.33                       \\

\multicolumn{1}{|l|}{\textbf{}} & & SPEC~\cite{paperSPEC_2007} &  ---  & 45.32                      \\
\cline{2-5}
\multicolumn{1}{|l|}{\textbf{}} & \multirow{2}{*}{\textbf{Late Fusion}} & USRF~\cite{paperUSRAF} &  --- & 39.75                      \\
\multicolumn{1}{|l|}{\textbf{}} & & \cellcolor{lightgray} \textbf{HRSF (ours)} & \cellcolor{lightgray} --- & \cellcolor{lightgray} \textbf{54.09}                       \\
\hline
\end{tabular}}
\end{table}

\begin{figure}[!ht]
    \centering
    \includegraphics[width=.54\textwidth]{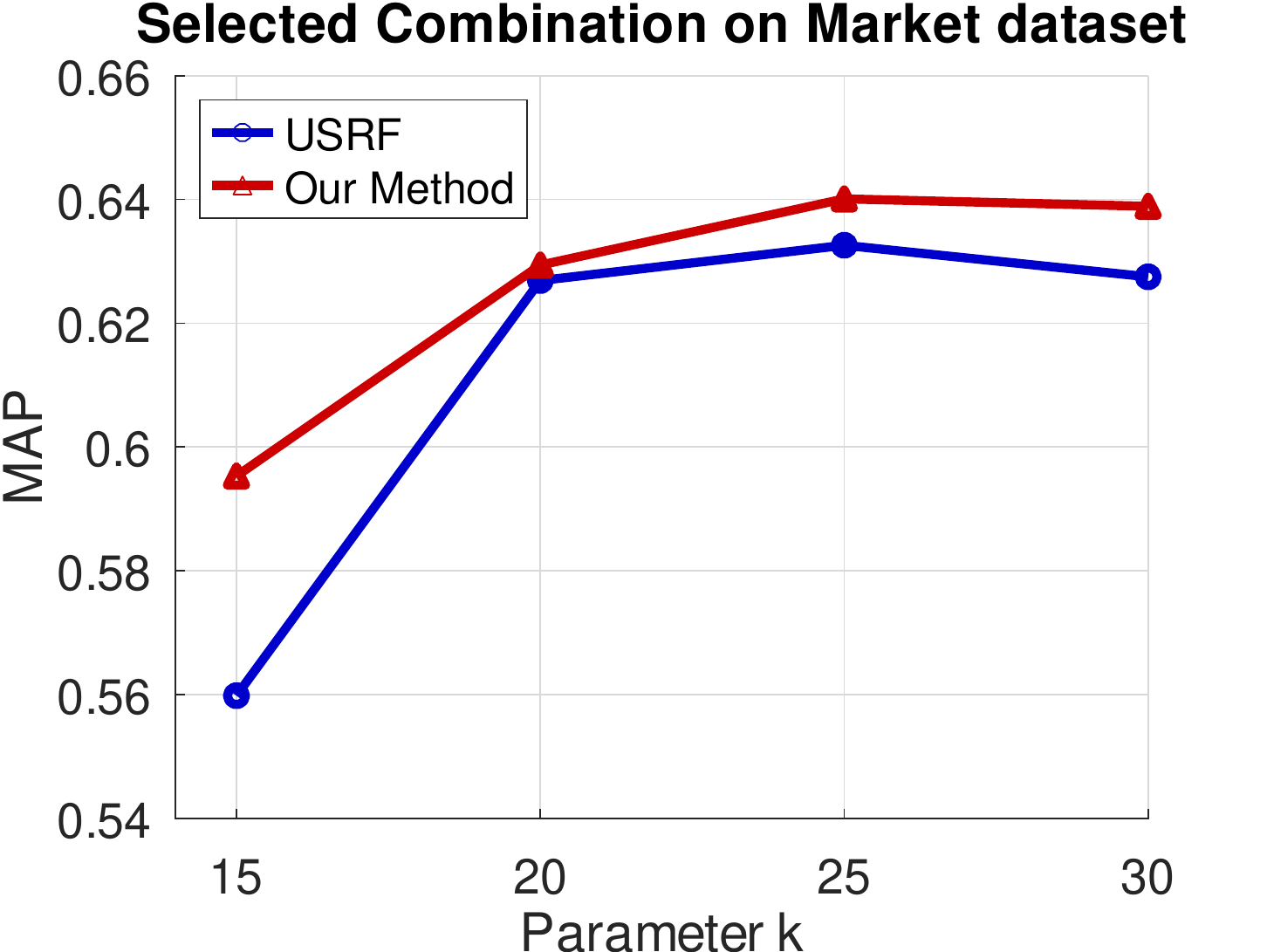}
    \caption{Selected Combination (among top-5) on Market considering MAP.}
    \label{fig:map_market}
\end{figure}

\begin{figure}[!ht]
    \centering
    \includegraphics[width=.54\textwidth]{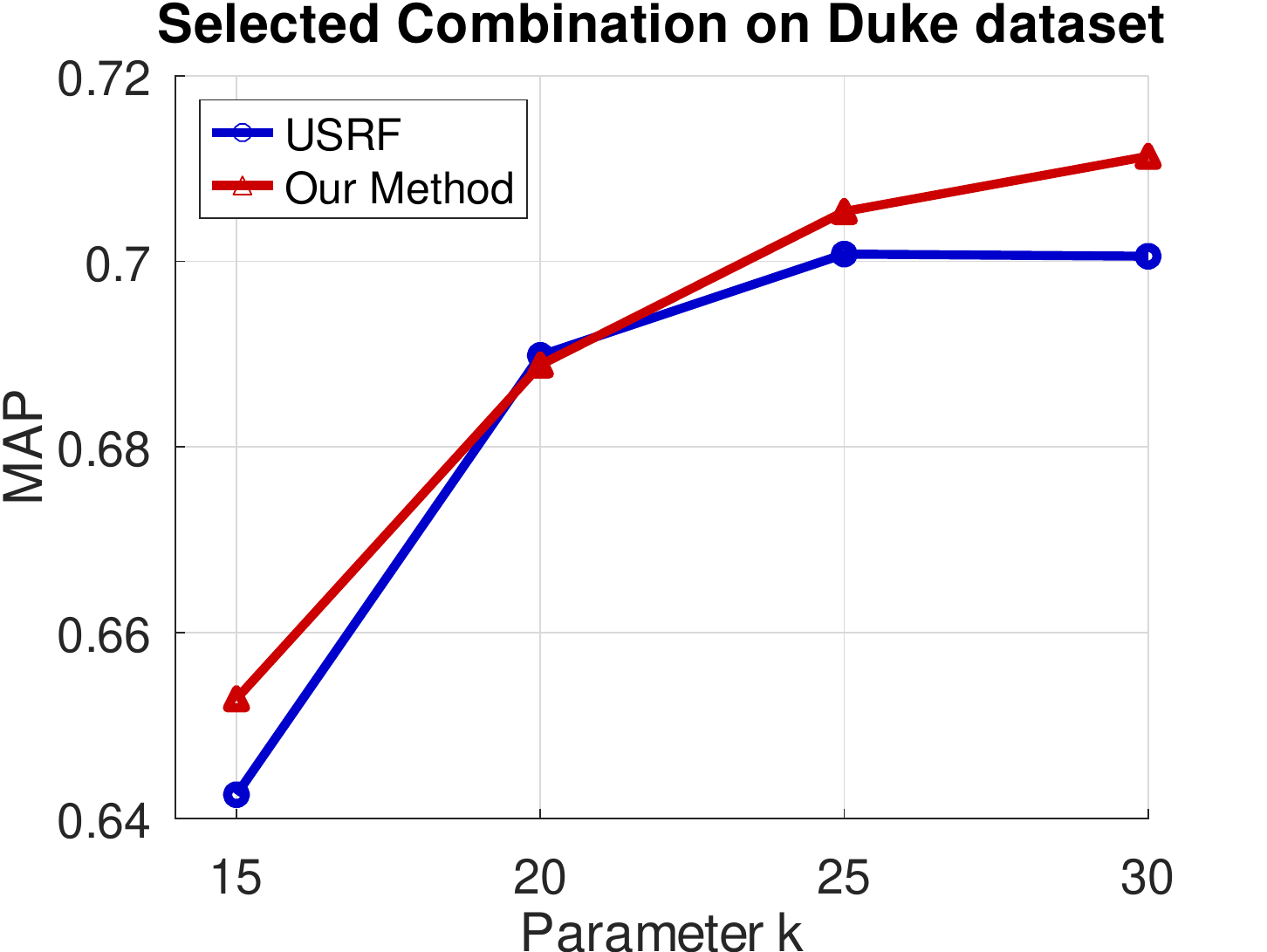}
    \caption{Selected Combination (among top-5) on Duke considering MAP.}
    \label{fig:map_duke}
\end{figure}

An experiment was conducted with the objective of clarifying the robustness of HRSF to different values of $k$ when compared to USRF.
Figures~\ref{fig:map_market} and~\ref{fig:map_duke} present the MAP of the best combination ($\mathfrak{X}^{*}$) among top-5 for different values of $k$ on Market and Duke datasets, respectively.
The MAP for $k=20$ is the same of Table~\ref{tab:fusion_baselines}.
Notice that, while both methods seem comparable for $k=20$, \textbf{our method provided a significant higher MAP for other values of $k$.} This is fundamental for unsupervised scenarios, where the optimal $k$ can be challenging to define.

\subsection{State-of-the-Art}

This section presents comparisons with recent baselines.
The taxonomy often varies in the literature; some of the subcategories of unsupervised Re-ID methods are:
(\textit{i}) \textbf{Cross-domain Methods:} they are trained on one (single-source) or more (multi-source) labeled source domains. The evaluation is conducted on the target dataset. It can be understood as a type of transfer learning;
(\textit{ii}) \textbf{Domain Adaptive Methods:} they use cross-domain training and perform a second step (adaptation) using unlabeled data from the target domain.

However, since they are all unsupervised and often there is overlap among the categories, we insert all of them into a single group named Unsupervised Domain Adaptation Methods~\cite{paperUDA}.
Table~\ref{tab:state_of_art} presents our results compared to around \textbf{20 state-of-the-art Unsupervised Domain Adaptation Methods~\cite{paperUDA}}.

In order to perform a fair comparison, it contains \textbf{only methods that did not use the labels from the target dataset for training} (train on CUHK03 and test on CUHK03 is not used, for example).
Therefore, supervised and semi-supervised methods are not included.
The abbreviations in parentheses indicate the datasets used for training~\footnote{C02 = CUHK02, C03 = CUHK03, M = Market1501, D = DukeMTMC, MT = MSMT17}.
For example, the use of (D, M) indicates that the reported result corresponds to a training done either on Duke or on Market dataset. The results reported on Market were trained on Duke and the results reported on Duke were trained on Market.
\textbf{None of the presented methods were trained using labels from the target dataset.}
The abbreviations were omitted for baselines that used more than 5 datasets as source for training (CAMEL~\cite{paperCAMEL} and baseline by~\cite{paperLargestMS}), but they can be consulted in~\cite{paperLargestMS}, which used similar baselines and protocol.

We provided the best results for each method (considering the original papers) with the objective of keeping the evaluation as far as possible.
Since the code and implementation are not available for the majority of methods, it is not possible to provide results considering the same sources in all cases.

Both MAP and R1 are reported in all the cases and our best results are presented in bold.
The baselines do not perform any form of selection.
We highlight that our method receives all the rankers as input and performs a wide selection among rankers with high and low effectiveness, which is a very challenging scenario.
In contrast, \textbf{none of the baselines are required to perform any selection} and the features are chosen manually.
\textbf{The selection stage is an important aspect and contribution of HRSF} that it is hard to replicate in the baselines.
Notice that our approach achieved competitive or superior results for all the evaluated datasets.

With the objective to facilitate the visualization of the best results in the state-of-the-art, Table~\ref{tab:state_of_art_rank} presents the rank of the methods according to MAP and R1.
The gray cells with bold values correspond to methods that have achieved a higher rank than HRSF in each case.
Our method achieved the best MAP on DukeMTMC and had the second position in the other two datasets.
For R1, HRSF is positioned among the top-4 in all cases.
The mean of the rank on each dataset is presented in the most right columns.
Notice that our method achieved one of the highest rank means among all of the methods, being only slightly behind ISSDA~\cite{paperISSDA}.
The comparisons show that, besides our results being among the best for all datasets, in some cases, other non-fusion based methods provided higher values than our approach.
There are some possible explanations for this:
\begin{itemize}
    \item Each dataset has different aspects (e.g., image resolution, picture angles, environment, number of images per person, dataset size). For this reason, \textbf{different methods may perform better or worse on distinct datasets};
    \item The Baseline by~\cite{paperLargestMS} performs a multi-source training.
    The results reported by~\cite{paperLargestMS} are based on the transfer learning of a \textbf{training performed on 7 Re-ID datasets}. It considers, by far, the \textbf{largest labeled source of all the baselines}, which leads to high results, especially on Market where it is ranked as the 2nd/3rd best R1/MAP (shown in Table~\ref{tab:state_of_art_rank});
    \item An idea that is exploited by some baselines is the generation of pseudo-labels. The most promising example is ISSDA~\cite{paperISSDA}, which has the best results on Market and is well ranked on DukeMTMC.
    Different from the others, ISSDA employs a self-supervised iterative pseudo-label generation and training.
    However, besides the effectiveness, \textbf{the authors~\cite{paperISSDA} claim that the training stage is very time-consuming} since it requires, among other aspects, the execution of a clustering algorithm.
    Furthermore, ISSDA has an average ranking of 1.5 against 1.67 of our method (Table~\ref{tab:state_of_art_rank}, MAP measure);
    \textbf{However, ISSDA does not report results on CUHK03. Our average ranking without considering CUHK03 is also 1.5};
    \item The MAR~\cite{paperMAR} performs \textbf{soft label generation}. The applied strategy is capable of achieving promising results for improving the DukeMTMC dataset with R1 measure. However, apparently, the quality of the soft labels varies according to the dataset.
\end{itemize}

Since the proposed method is flexible, among future works, we intend to expand our approach by exploiting the use of pseudo-labels.
This can be done by modifying HRSF or by providing methods like ISSDA~\cite{paperISSDA} and MAR~\cite{paperMAR} as input, for example.
In this work, only the rankers in Table~\ref{tab:descriptors} were used.

\begin{table}[H]
\centering
\caption{State-of-the-art comparison considering MAP (\%) and R-01 (\%).}
\label{tab:state_of_art}
\resizebox{.75\textwidth}{!}{ 
\begin{tabular}{l|c|c||c|c||c|c|}
\cline{2-7}
                                              & \multicolumn{6}{c|}{\textbf{Datasets}}                                                                                                                                                                             \\ \cline{2-7} 
                                              & \multicolumn{2}{c||}{\textbf{Market1501}}                             & \multicolumn{2}{c||}{\textbf{DukeMTMC}}                               & \multicolumn{2}{c|}{\textbf{CUHK03}}                                 \\ \cline{2-7}
       & \multicolumn{1}{c|}{\textbf{R1}} & \multicolumn{1}{c||}{\textbf{MAP}} & \multicolumn{1}{c|}{\textbf{R1}} & \multicolumn{1}{c||}{\textbf{MAP}} & \multicolumn{1}{c|}{\textbf{R1}} & \multicolumn{1}{c|}{\textbf{MAP}} \\ \cline{2-7}
\hline
\multicolumn{7}{|c|}{\textbf{Unsupervised Domain Adaptation Methods}} \\ \hline
\hline
\multicolumn{1}{|l|}{ARN~\cite{paperARN}}            & 70.3                             & 39.4                              & 60.2                             & 33.4                              & ---                              & ---                               \\ \hline
\multicolumn{1}{|l|}{EANet~\cite{paperEANET}}          & 66.4                             & 40.6                              & 45.0                             & 26.4                              & 51.4                             & 31.7                              \\ \hline
\multicolumn{1}{|l|}{ECN~\cite{paperECN}}            & 75.1                             & 43.0                              & 63.3                             & 40.4                              & ---                              & ---                               \\ \hline
\multicolumn{1}{|l|}{MAR~\cite{paperMAR}}            & 67.7                             & 40.0                              & 87.1                             & 48.0                              & ---                              & ---                               \\ \hline
\multicolumn{1}{|l|}{TAUDL~\cite{paperTAUDL}}          & 63.7                             & 41.2                              & 61.7                             & 43.5                              & 44.7                             & 31.2                              \\ \hline
\multicolumn{1}{|l|}{UTAL~\cite{paperUTAL}}           & 69.2                             & 46.2                              & 62.3                             & 44.6                              & 56.3                             & 42.3                              \\ \hline
\multicolumn{1}{|l|}{HHL (D,M)~\cite{paperHHL}}      & 62.2                             & 31.4                              & 46.9                             & 27.2                              & ---                              & ---                               \\ \hline
\multicolumn{1}{|l|}{HHL (C03)~\cite{paperHHL}}      & 56.8                             & 29.8                              & 42.7                             & 23.4                              & ---                              & ---                               \\ \hline
\multicolumn{1}{|l|}{ATNet (D,M)~\cite{paperATNET}}    & 55.7                             & 25.6                              & 45.1                             & 24.9                              & ---                              & ---                               \\ \hline
\multicolumn{1}{|l|}{CSGLP (D,M)~\cite{paperCSGLP}}    & 63.7                             & 33.9                              & 56.1                             & 36.0                              & ---                              & ---                               \\ \hline
\multicolumn{1}{|l|}{ISSDA (D,M)~\cite{paperISSDA}}    & 81.3                             & 63.1                              & 72.8                             & 54.1                              & ---                              & ---                               \\ \hline
\multicolumn{1}{|l|}{EANet (C03)~\cite{paperEANET}}    & 59.4                             & 33.3                              & 39.3                             & 22.0                              & ---                              & ---                               \\ \hline
\multicolumn{1}{|l|}{EANet (D,M)~\cite{paperEANET}}    & 61.7                             & 32.9                              & 51.4                             & 31.7                              & ---                              & ---                               \\ \hline
\multicolumn{1}{|l|}{SPGAN (D,M)~\cite{paperSPGAN}}    & 43.1                             & 17.0                              & 33.1                             & 16.7                              & ---                              & ---                               \\ \hline
\multicolumn{1}{|l|}{DAAM (D,M)~\cite{paperDAAM}}     & 42.3                             & 17.5                              & 29.3                             & 14.5                              & ---                              & ---                               \\ \hline
\multicolumn{1}{|l|}{AF3 (D,M)~\cite{paperAF3}}      & 67.2                             & 36.3                              & 56.8                             & 37.4                              & ---                              & ---                               \\ \hline
\multicolumn{1}{|l|}{AF3 (MT)~\cite{paperAF3}}       & 68.0                             & 37.7                              & 66.3                             & 46.2                              & ---                              & ---                               \\ \hline
\multicolumn{1}{|l|}{PAUL (MT)~\cite{paperPAUL}}      & 68.5                             & 40.1                              & 72.0                             & 53.2                              & ---                              & ---                               \\ \hline
\multicolumn{1}{|l|}{EMTL (C02+D+M)~\cite{paperEMTL}} & 52.8                             & 25.1                              & 39.7                             & 22.3                              & ---                              & ---                               \\ \hline
\multicolumn{1}{|l|}{CAMEL~\cite{paperCAMEL}}          & 54.5                             & 26.3                              & ---                              & ---                               & 31.9                             & ---                               \\ \hline
\multicolumn{1}{|l|}{Baseline by~\cite{paperLargestMS}}    & 80.5                             & 56.8                              & 67.4                             & 46.9                              & 29.4                             & 27.4                              \\ \hline
\hline
\multicolumn{7}{|c|}{\textbf{Unsupervised Selection and Fusion (ours)}} \\ \hline
\hline
\multicolumn{1}{|l|}{\textbf{HRSF ($\mathfrak{X}^{*}_{2}$)}} & 74.32 &  60.89 & 76.80  & 68.51 & \textbf{39.04} & \textbf{39.69} \\ \hline
\multicolumn{1}{|l|}{\textbf{HRSF ($\mathfrak{X}^{*}_{3}$)}} & 75.56 & 62.64 & \textbf{77.24}  & \textbf{68.88} & 39.13 & 39.58 \\ \hline
\multicolumn{1}{|l|}{\textbf{HRSF ($\mathfrak{X}^{*}_{4}$)}} & \textbf{75.71} &  \textbf{62.94} & 76.89  & 68.56 & 38.02 & 38.80 \\ \hline
\multicolumn{1}{|l|}{\textbf{HRSF ($\mathfrak{X}^{*}_{5}$)}} & 74.00 &  60.69 & 76.39 & 67.72 & 36.15 & 37.11 \\ \hline
\multicolumn{1}{|l|}{\textbf{HRSF ($\mathfrak{X}^{*}_{6}$)}} & 73.57 &  59.85 & 75.90  & 66.96 & 35.46 & 36.17 \\ \hline
\multicolumn{1}{|l|}{\textbf{HRSF ($\mathfrak{X}^{*}$, best result)}} & \textbf{75.71} &  \textbf{62.94} & \textbf{77.24}  & \textbf{68.88} & \textbf{39.04} & \textbf{39.69} \\ \hline
\end{tabular}
}
\end{table}

\begin{table}[H]
\centering
\caption{State-of-the-art methods ranked by their results.}
\label{tab:state_of_art_rank}
\resizebox{.93\textwidth}{!}{ 
\begin{tabular}{l|c|c||c|c||c|c||c|c|}
\cline{2-9} 
                                              & \multicolumn{2}{c||}{\textbf{Market1501}}                             & \multicolumn{2}{c||}{\textbf{DukeMTMC}}                               & \multicolumn{2}{c||}{\textbf{CUHK03}}   & \multicolumn{2}{c|}{\textbf{Mean}}                              \\ \cline{2-9}
       & \multicolumn{1}{c|}{\textbf{R1}} & \multicolumn{1}{c||}{\textbf{MAP}} & \multicolumn{1}{c|}{\textbf{R1}} & \multicolumn{1}{c||}{\textbf{MAP}} & \multicolumn{1}{c|}{\textbf{R1}} & \multicolumn{1}{c||}{\textbf{MAP}}  & \multicolumn{1}{c|}{\textbf{R1}} & \multicolumn{1}{c|}{\textbf{MAP}} \\
\cline{2-9}
\hline
\multicolumn{9}{|c|}{\textbf{Unsupervised Domain Adaptation Methods}} \\ \hline
\hline
\multicolumn{1}{|l|}{ARN~\cite{paperARN}}            & 5                             & 10                              & 10                             & 12                              & ---                              & ---  & 7.5 & 11                            \\ \hline
\multicolumn{1}{|l|}{EANet~\cite{paperEANET}}          & 11                             & 7                              & 16                             & 15                              & \cellcolor{lightgray}\textbf{2}                             & 3    & 9.67 &  8.34                        \\ \hline
\multicolumn{1}{|l|}{ECN~\cite{paperECN}}            & 4                             & 5                              & 7                             & 9                              & ---                              & ---    & 5.5 &  7                          \\ \hline
\multicolumn{1}{|l|}{MAR~\cite{paperMAR}}            & 9                             & 9                             & \cellcolor{lightgray}\textbf{1}                             & 4                              & ---                              & ---     & 5 & 6.5                          \\ \hline
\multicolumn{1}{|l|}{TAUDL~\cite{paperTAUDL}}          & 12                             & 6                              & 9                             & 8                             & \cellcolor{lightgray}\textbf{3}                             & 4      & 8 & 6                        \\ \hline
\multicolumn{1}{|l|}{UTAL~\cite{paperUTAL}}           & 6                             & 4                              & 8                             & 7                              & \cellcolor{lightgray}\textbf{1}                            & \cellcolor{lightgray}\textbf{1}       & 5 &  4                      \\ \hline
\multicolumn{1}{|l|}{HHL (D,M)~\cite{paperHHL}}      & 14                             & 16                             & 14                             & 14                              & ---                              & ---      & 14 & 15                         \\ \hline
\multicolumn{1}{|l|}{HHL (C03)~\cite{paperHHL}}      & 17                             & 17                              & 17                             & 17                              & ---                              & ---     & 17 &  17                         \\ \hline
\multicolumn{1}{|l|}{ATNet (D,M)~\cite{paperATNET}}    & 18                             & 19                             & 15                             & 16                              & ---                              & ---   & 16.5 & 17.5                            \\ \hline
\multicolumn{1}{|l|}{CSGLP (D,M)~\cite{paperCSGLP}}    & 13                             & 13                              & 12                            & 11                              & ---                              & ---    & 12.5 &    12                        \\ \hline
\multicolumn{1}{|l|}{ISSDA (D,M)~\cite{paperISSDA}}    & \cellcolor{lightgray}\textbf{1}                             & \cellcolor{lightgray}\textbf{1}                              & 3                             & 2                              & ---                              & ---      & \cellcolor{lightgray}\textbf{2} &    \cellcolor{lightgray}\textbf{1.5}                      \\ \hline
\multicolumn{1}{|l|}{EANet (C03)~\cite{paperEANET}}    & 16                             & 14                              & 19                             & 19                              & ---                              & ---      & 11.67 & 16.5                          \\ \hline
\multicolumn{1}{|l|}{EANet (D,M)~\cite{paperEANET}}    & 15                             & 15                              & 13                         & 13                             & ---                              & ---          & 14 & 14                      \\ \hline
\multicolumn{1}{|l|}{SPGAN (D,M)~\cite{paperSPGAN}}    & 21                             & 22                              & 20                             & 20                              & ---                              & ---      & 20.5 & 21                         \\ \hline
\multicolumn{1}{|l|}{DAAM (D,M)~\cite{paperDAAM}}     & 22                             & 21                              & 21                             & 21                              & ---                              & ---        & 21.5 & 21                       \\ \hline
\multicolumn{1}{|l|}{AF3 (D,M)~\cite{paperAF3}}      & 10                             & 12                              & 11                             & 10                              & ---                              & ---        & 10.5 & 11                        \\ \hline
\multicolumn{1}{|l|}{AF3 (MT)~\cite{paperAF3}}       & 8                             & 11                              & 6                             & 6                              & ---                              & ---           & 7 & 8.5                     \\ \hline
\multicolumn{1}{|l|}{PAUL (MT)~\cite{paperPAUL}}      & 7                             & 8                              & 4                            & 3                              & ---                              & ---           & 5.5 &  5.5                   \\ \hline
\multicolumn{1}{|l|}{EMTL (C02+D+M)~\cite{paperEMTL}} & 20                             & 20                             & 18                             & 18                              & ---                              & ---          & 19 & 19                     \\ \hline
\multicolumn{1}{|l|}{CAMEL~\cite{paperCAMEL}}          & 19                             & 18                             & ---                              & ---                              & 5                             & ---       & 12 &  18                       \\ \hline
\multicolumn{1}{|l|}{Baseline by~\cite{paperLargestMS}}    & \textbf{2}                             & 3                              & 5                             & 5                              & 6                            & 5            & 4.34 & 4.34                   \\ \hline
\hline
\multicolumn{9}{|c|}{\textbf{Unsupervised Selection and Fusion (ours)}} \\ \hline
\hline
\multicolumn{1}{|l|}{\cellcolor{lightgray}\textbf{HRSF ($\mathfrak{X}^{*}$, best result)}} & \cellcolor{lightgray}\textbf{3} &  \cellcolor{lightgray}\textbf{2} & \cellcolor{lightgray}\textbf{2}  & \cellcolor{lightgray}\textbf{1} & \cellcolor{lightgray}\textbf{4} & \cellcolor{lightgray}\textbf{2}  & \cellcolor{lightgray}\textbf{3} &   \cellcolor{lightgray}\textbf{1.67} \\ \hline
\end{tabular}
}
\end{table}

\subsection{Visual Results}

\begin{figure*}[ht!]
    \centering
    \includegraphics[width=.91\textwidth]{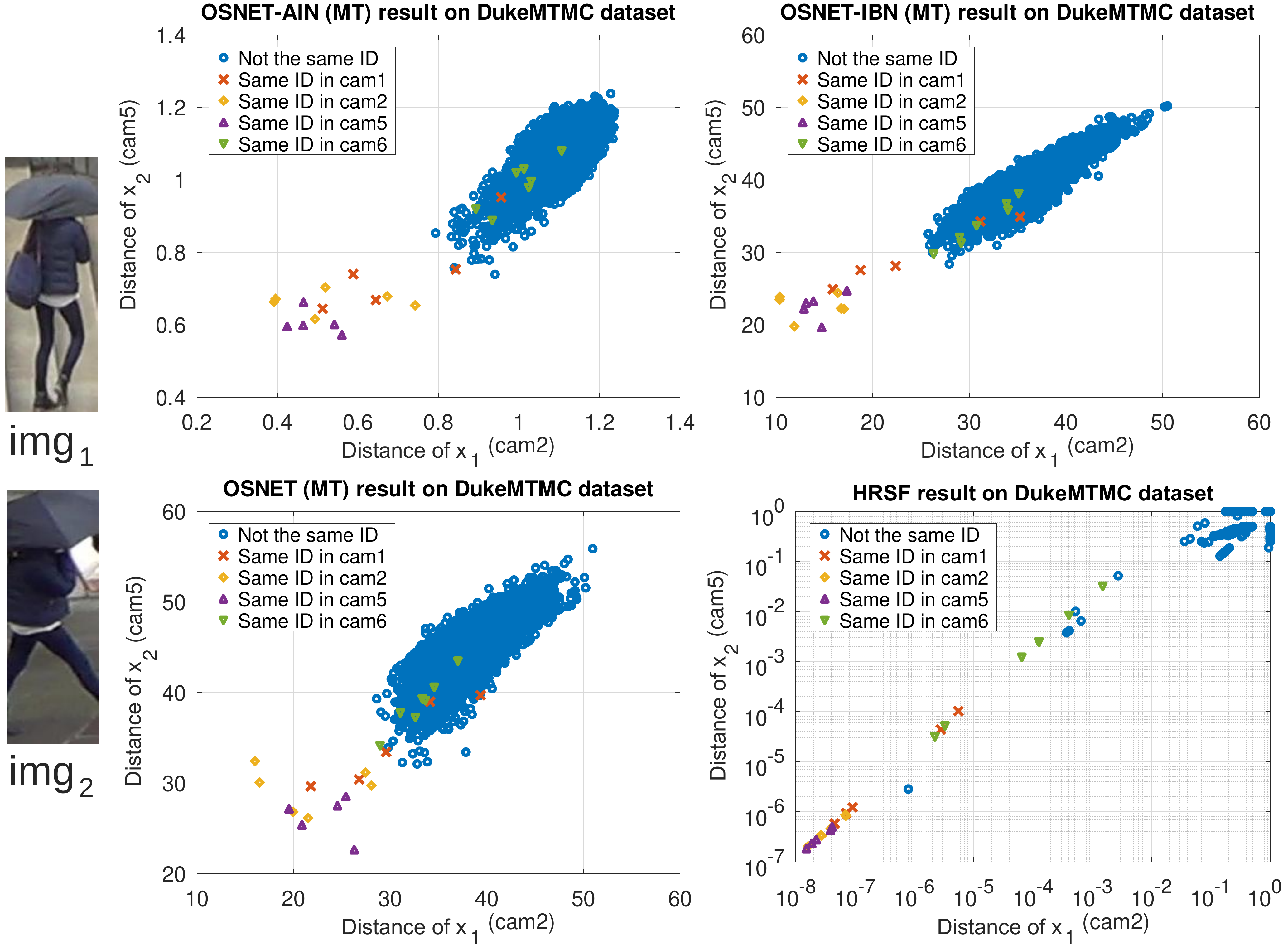}
    \vspace{-3mm}
    \caption{Distance distribution for two query images on DukeMTMC dataset.}
    \label{fig:comparison_queries}
    \vspace{-2mm}
\end{figure*}

Some qualitative results were also elaborated to evince the quality of our obtained results.
Two different queries ($x_1$, $x_2$) for the same person (ID) were chosen from the DukeMTMC dataset.
Figure~\ref{fig:comparison_queries} presents a graph for each ranker that composes the best combination ($\mathfrak{X}^{*}$) on DukeMTMC and the HRSF result.
Each dot represents a gallery image, which is positioned in the graph according to its distance to the query images ($x_1$, $x_2$).
The idea is that, since the query images are of the same person, the distance between them should be small and the images of the same ID should be closer to the bottom left corner.
Images obtained from different camera views are presented in different symbols.
It shows that the HRSF method was capable of reducing the distance of all the images belonging to the same class when compared to isolated rankers (OSNET, OSNET-IBN, OSNET-AIN).

\begin{figure}[H]
    \centering
    \begin{tabular}{c}
        \textbf{{\normalsize OSNET-AIN (MT)}} \\ \includegraphics[width=.96\textwidth]{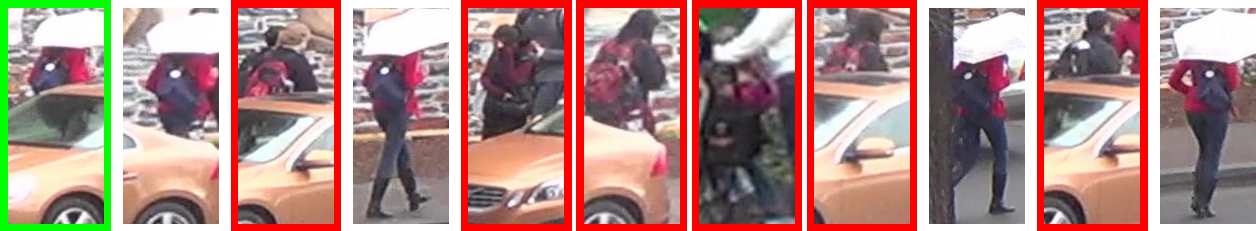} \\
        \textbf{{\normalsize OSNET-IBN (MT)}} \\ \includegraphics[width=.96\textwidth]{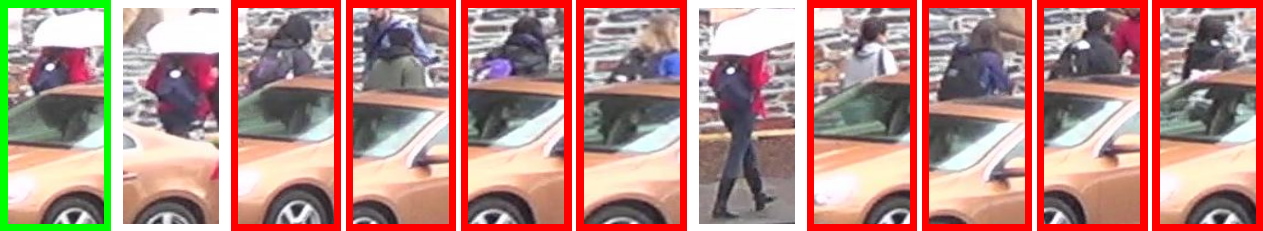} \\
        \textbf{{\normalsize OSNET (MT)}} \\ \includegraphics[width=.96\textwidth]{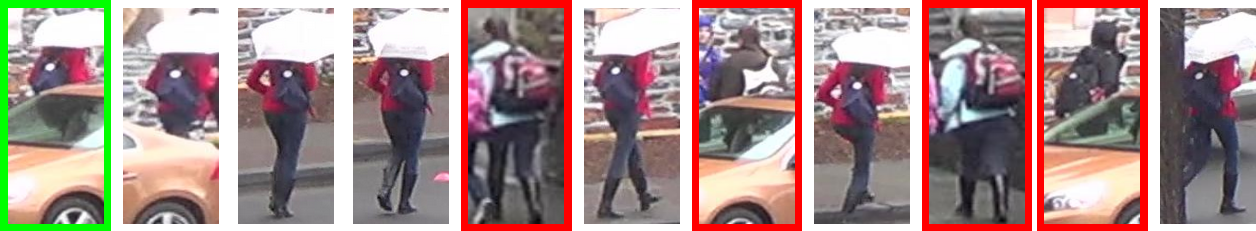} \\
        \textbf{{\normalsize HRSF Fusion ($\mathfrak{X}^{*}$)}} \\ \includegraphics[width=.96\textwidth]{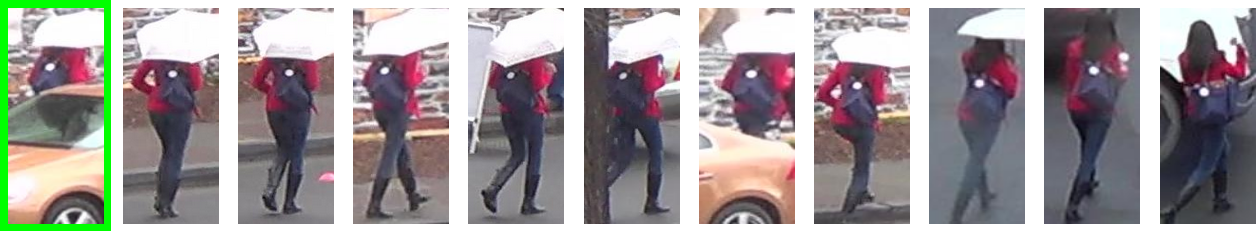} \\
    \end{tabular}
    \caption{Examples to illustrate the impact of HRSF selection and fusion on DukeMTMC dataset.}
    \label{fig:visual_result_duke}
\end{figure}

\begin{figure}[H]
    \centering
    \begin{tabular}{c}
        \textbf{{\normalsize OSNET-AIN (MT)}} \\ \includegraphics[width=.96\textwidth]{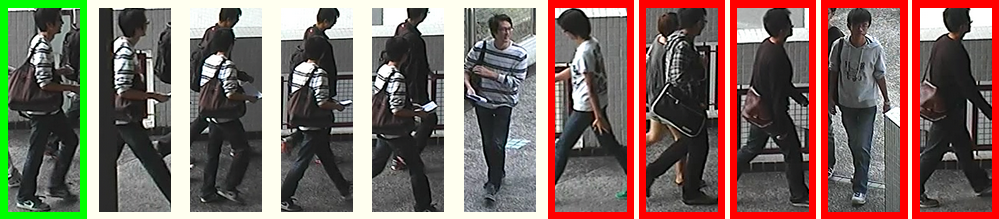} \\
        \textbf{{\normalsize OSNET-IBN (MT)}} \\ \includegraphics[width=.96\textwidth]{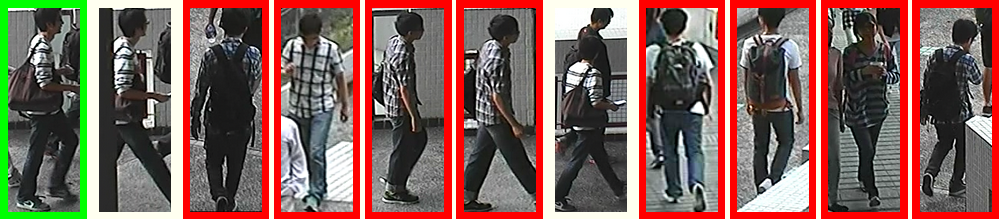} \\
        \textbf{{\normalsize HRSF Fusion ($\mathfrak{X}^{*}$)}} \\ \includegraphics[width=.96\textwidth]{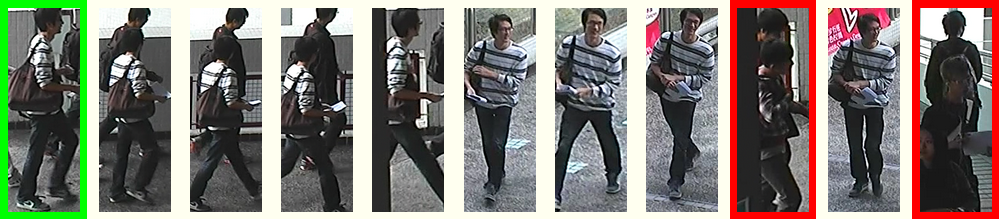} \\
    \end{tabular}
    \caption{Examples to illustrate the impact of HRSF selection and fusion on CUHK03 dataset.}
    \label{fig:visual_result_cuhk03}
\end{figure}

Figures~\ref{fig:visual_result_duke} and~\ref{fig:visual_result_cuhk03} present examples of visual queries on the DukeMTMC and CUHK03 datasets, respectively.
The results are shown for the best combination obtained by HRSF ($\mathfrak{X}^{*}$) and the rankers that compose it.
The query image is presented with green borders, and the wrong results with red borders.
Notice that, in these cases, beyond selecting the best results, our approach was also capable of removing most of the incorrectly retrieved images.

\section{Conclusion} 
\label{secConc}

In this work, we have presented an approach to select and aggregate results from different Re-ID methods without using the target dataset training labels.
The relationship among the dataset elements is modeled using hypergraph structures both for query performance prediction and fusion.
In most of the cases, the best rankers were properly selected and fused.
Our approach achieved significant gains in the majority of the scenarios and results competitive or superior to the state-of-the-art, including the most recent methods.

As future work, we intend to investigate the use of the proposed approach in multi-query scenarios and further expand the experimental protocol, including other datasets and feature extractors. 
Another idea is to expand our approach by exploiting the use of pseudo-labels.
Our method, as currently presented, is ready to be executed in scenarios that do not require real time.
In addition,  we intend to address these real time scenarios in future work, where optimizations and parallelization techniques can be employed.

\section*{\uppercase{Acknowledgements}}
The authors are grateful to the S\~{a}o Paulo Research Foundation - FAPESP (grant \#2018/15597-6 and \#2017/25908-6), Brazilian National Council for Scientific and Technological Development - CNPq (grant \#308194/2017-9), Microsoft Research, and Petrobras (grant \#2017/00285-6).

\small
\bibliographystyle{elsarticle-num} 
\bibliography{reid_bib}

\end{document}